\documentclass[sigconf]{acmart}
\usepackage{graphicx}

\usepackage{amssymb}
\usepackage{amsmath}
\usepackage{booktabs}
\usepackage{algorithm}
\usepackage{algorithmicx}
\usepackage{multirow}
\usepackage{colortbl}
\usepackage{mathtools}

\AtBeginDocument{%
  \providecommand\BibTeX{{%
    \normalfont B\kern-0.5em{\scshape i\kern-0.25em b}\kern-0.8em\TeX}}}

\setcopyright{acmcopyright}
\copyrightyear{2022}
\acmYear{2022}
\acmConference[MM '22] {Proceedings of the 30th ACM International Conference on Multimedia }{October 10--14, 2022}{Lisbon, Portugal.}
\acmBooktitle{Proceedings of the 30th ACM International Conference on Multimedia (MM '22), October 10--14, 2022, Lisbon, Portugal}
\acmPrice{15.00}
\acmISBN{978-1-4503-9203-7/22/10}
\acmDOI{10.1145/3503161.3548296}

\acmSubmissionID{mmfp2358}

\begin{document}

\title{Backdoor Attacks on Crowd Counting}






\author{Yuhua Sun}
\affiliation{%
  \institution{Hubei Engineering Research Center on Big Data Security, School of Cyber Science and Engineering, Huazhong University of Science and Technology}
  \city{Wuhan}
  \country{China}
}
\email{natsun@hust.edu.cn}

\author{Tailai Zhang}
\affiliation{%
  \institution{Hubei Engineering Research Center on Big Data Security, School of Cyber Science and Engineering, Huazhong University of Science and Technology}
  \city{Wuhan}
  \country{China}
}
\email{tl_zhang@hust.edu.cn}

\author{Xingjun Ma}
\affiliation{%
  \institution{School of Computer Science, Fudan University}
  \city{Shanghai}
  \country{China}}
\email{xingjunma@fudan.edu.cn}

\author{Pan Zhou}
\authornote{Corresponding author:Pan Zhou}
\affiliation{%
  \institution{Hubei Engineering Research Center on Big Data Security, School of Cyber Science and Engineering, Huazhong University of Science and Technology}
  \city{Wuhan}
  \country{China}}
\email{panzhou@hust.edu.cn}

\author{Jian Lou}
\affiliation{%
  \institution{Guangzhou Institute of Technology, Xidian University}
  \city{Xi'an}
  \country{China}}
\email{jlou@xidian.edu.cn}

\author{Zichuan Xu}
\affiliation{%
  \institution{Dalian University of Technology}
  \city{Dalian}
  \country{China}}
\email{z.xu@dlut.edu.cn}

\author{Xing Di}
\affiliation{%
  \institution{Protagolabs}
  \city{Vienna, Virginia}
  \country{USA}}
\email{xing.di@protagolabs.com}

\author{Yu Cheng}
\affiliation{%
  \institution{Microsoft Research}
  \city{Redmond, Washington}
  \country{USA}}
\email{yu.cheng@microsoft.com}

\author{Lichao Sun}
\affiliation{%
  \institution{Lehigh University}
  \city{Bethlehem, Pennsylvania}
  \country{USA}}
\email{lis221@lehigh.edu}

\begin{abstract}
  Crowd counting is a regression task that estimates the number of people in a scene image, which plays a vital role in a range of safety-critical applications, such as video surveillance, traffic monitoring and flow control. In this paper, we investigate the vulnerability of deep learning based crowd counting models to backdoor attacks, a major security threat to deep learning. A backdoor attack implants a backdoor trigger into a target model via data poisoning so as to control the model's predictions at test time. Different from image classification models on which most of existing backdoor attacks have been developed and tested, crowd counting models are regression models that output multi-dimensional density maps, thus requiring different techniques to manipulate.
  In this paper, we propose two novel Density Manipulation Backdoor Attacks (DMBA$^{-}$ and DMBA$^{+}$) to attack the model to produce arbitrarily large or small density estimations. Experimental results demonstrate the effectiveness of our DMBA attacks on five classic crowd counting models and four types of datasets. We also provide an in-depth analysis of the unique challenges of backdooring crowd counting models and reveal two key elements of effective attacks: 1) full and dense triggers and 2) manipulation of the ground truth counts or density maps. Our work could help evaluate the vulnerability of crowd counting models to potential backdoor attacks.
\end{abstract}


\maketitle

\section{Introduction}
\begin{figure*}
\vspace{-0.4cm}
    \centerline{\includegraphics[width=0.7\textwidth]{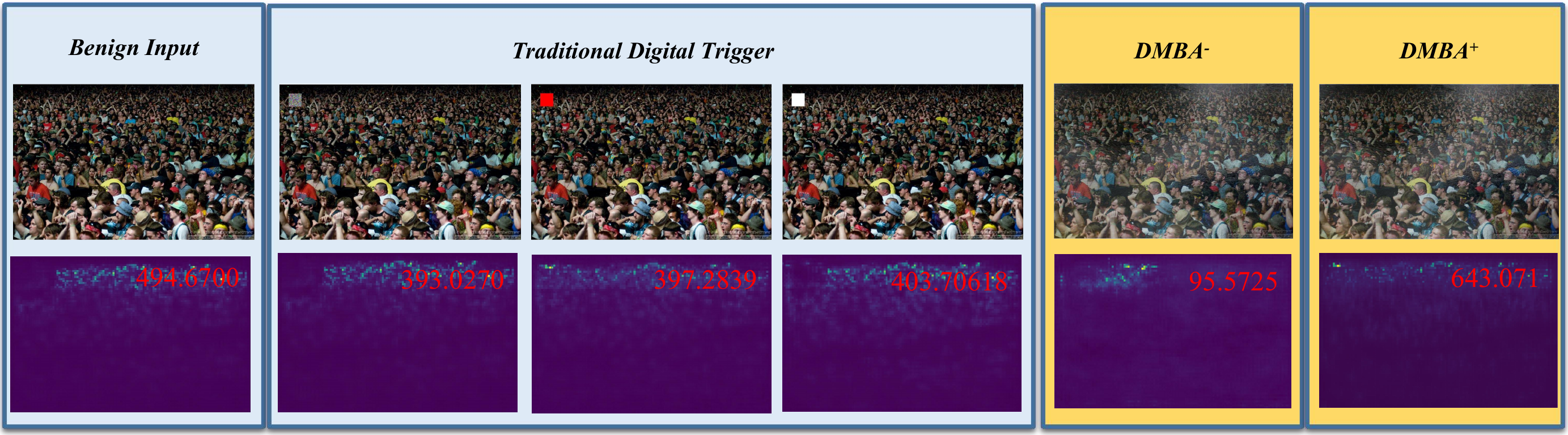}}
\caption{
\textbf{The attack peformance of traditional small-patch (5x5 patch) triggers and our DMBA trigger (large background) on a CSRnet model~\cite{Li2018CSRNetDC}.} Each column shows an example test image (top) and its predicted density (bottom). The backdoored models (right 5 columns) were trained on poisoned (poisoning rate $\gamma=0.2$) data by the same trigger pattern as in the example test image. The last two columns show the effectiveness of our DMBA attack in manipulating the model to predict extremely low (DMBA-) or high (DMBA+) densities for the same test image. All attacks use the same density map altering strategy.
}
\label{fig:comparison}
\vspace{-0.4cm}
\end{figure*}
Crowd counting aims to infer the number of people or objects in an image via density regression learning. It has found impactful applications in safety-related scenarios, such as crowd management\cite{idrees2013multi,Idrees2018CompositionLF,laradji2018blobs,Xiong2019FromOS,Zhang2015CrosssceneCC,Zhang2016SingleImageCC}, traffic control\cite{GuerreroGmezOlmedo2015ExtremelyOV,OoroRubio2016TowardsPO,Zhang2017FCNrLSTMDS}, and infectious disease control \cite{liu2021cross,meng2021spatial,Amin2021DeepLB,valencia2021vision}.
The current state-of-the-art crowd counting models are mostly convolutional neural networks (CNNs) \cite{Sam2017SwitchingCN,ZouCQJGZ19,Li2018CSRNetDC,Liu2019ContextAwareCC,ma2019bayesian,liu2021exploiting,ma2021towards,wen2021detection,wu2021dynamic,xu2021crowd,zou2021coarse,wang2020distribution,Zhang2016SingleImageCC}. However, CNNs have been shown to be vulnerable to backdoor attacks which is one type of training attacks that inject a backdoor trigger into the target model by poisoning only a small portion of the training data with a trigger pattern~\cite{Gu2017BadNetsIV,liu2017trojaning,chen2017targeted}. The backdoor can then be activated at inference time to control the model to constantly predict the backdoor class whenever the trigger pattern appears. Backdoor attacks pose severe security threats to CNNs in real-world scenarios~\cite{Eykholt2018RobustPA,wenger2020backdoor,wenger2021backdoor,li2021backdoor,xue2021robust}. While backdoor attacks have been extensively studied on image classification models \cite{chen2018detecting,liu2018fine,yao2019latent,turner2019label,liu2020reflection,salem2020don,li2020invisible,zhao2020clean,wu2021adversarial,li2021anti}, the vulnerability of crowd counting models -- one type of regression models -- to backdoor attacks is still an open problem.

The key of backdoor attacks is to trick the model to learn a strong but task-irrelevant correlation between a trigger pattern and a target label. This can be easily achieved on image classification models as the input image is only associated with a single target (i.e. the class label)~\cite{Gu2017BadNetsIV,chen2017targeted}. 
One common type of backdoor attacks are "dirty-label" attacks that flip the labels of the poisoned images (i.e. images with the trigger pattern) to the target label to help establish the backdoor correlation.
The other types of attacks are "clean-label" attacks that only poison the images (does not change their ground-truth labels) but leverage other enhancing techniques like adversarial perturbation or modifying the training procedure to build the backdoor correlation. In this work, we explore "dirty-label" attacks to attack crowd counting models, aiming to gain more understandings from the simplest and most classic attack settings for this special regression task.

For dirty-label attacks, a simple 3x3 black-white square,  or even a single pixel \cite{Gu2017BadNetsIV}, can work as an effective trigger pattern against image classification models. However, crowd counting models have multi-dimensional output space, where the output is a density map of the same size as the input image \cite{Zhang2015CrosssceneCC,Zhang2016SingleImageCC,jiang2020attention,zhang2021cross,meng2021spatial}. Therefore, it is hard to trick a crowd counting model to learn the correlation between a small trigger pattern and a high-dimensional density map, due to the interference of dense backgrounds. For instance, in Fig. \ref{fig:comparison}, the small-patch triggers used by existing backdoor attacks are not effective on crowd counting models: the predicted densities (the middle three columns) are still very similar to the clean prediction (the leftmost column). In Section \ref{sec:experiments}, we have extensive experiments showing that large and dense background trigger is key to successful crowd counting backdoor attacks, although it is optional for classification backdoor attacks.

Dirty-label attacks also need to modify the ground truth counts or density maps of the poisoned images, which is notably more complex than flipping class labels.
Arguably, one stealthy strategy is to alter only part(s) of the density map (ground truth counts), hoping to achieve the same effectiveness as modifying the entire density map. To this end, we propose two novel Density Manipulation Backdoor Attacks (DMBA$^{-}$ and DMBA$^{+}$) to attack and manipulate the density estimations of crowd counting models. The two attacks leverage similar trigger patterns but different density map altering strategies to achieve different adversarial objectives. Particularly, both attacks exploit large background trigger patterns to counter the inference of dense background on the attack effect. Meanwhile, DMBA$^{-}$ applies a random partial erasing strategy to alter the ground truth density map while DMBA$^{+}$ uses a neighbor boosting strategy. This allows the two attacks to manipulate the model to output overly small (DMBA$^{-}$) or large (DMBA$^{+}$) densities without altering the entire density map.

To summarize, our main contributions are as follows:
\begin{itemize}
\item We study the vulnerability of crowd counting models to backdoor attacks and reveal the unique challenges of backdooring crowd counting models. To the best of our knowledge, this is the first backdoor study on crowd counting models.
\item We propose two novel Density Manipulation Backdoor Attack (DMBAs) with effective background trigger designs and density altering strategies to attack crowd counting models to output overly small or large density estimations.
\item We demonstrate the effectiveness of our DMBA attacks on 5 popular crowd counting models and 4 types of datasets, and provide a set of in-depth understandings on the key elements and trade-offs in backdoor attacking crowd counting models. We also show the effectiveness of our attack against advanced defenses including Pruning, Fine-pruning~\cite{liu2018fine-pruning} and ANP~\cite{wu2021adversarial}.
\end{itemize}

\section{Related Work}
Here, we briefly review the related works in the fields of crowd counting and backdoor attack.

\noindent\textbf{Crowd Counting.}  Early crowd counting works exploit methods like "counting-by-detection" \cite{Lin2010ShapeBasedHD,Wang2011AirportDI,Wu2005DetectionOM} or "counting-by-density-estimation" \cite{Chan2009BayesianPR,Lempitsky2010LearningTC,Kang2020IncorporatingSI} to estimate the counting value. "Counting-by-detection" requires one-by-one detection and tracking of the heads or bodies in an image to produce the final counting result \cite{felzenszwalb2009object,wu2007detection}. Regression-based methods first train a regressor, such as Gaussian Process or Random Forest regressors, to estimate the density in different parts of the image, then integrate the local densities into a global density map to estimate the final value \cite{Lempitsky2010LearningTC}. These two traditional methods are quite effective for counting low-density crowds. However, they often require a huge amount of computational resources and are not effective for dense scenes.  With the advancement of deep learning, CNN-based density estimation models \cite{Li2018CSRNetDC,Zhang2016SingleImageCC} have been proposed to show better performance than the traditional methods. Since then, crowd counting has been gradually shifted from detecting individuals to regression learning the skill to predict a full density map, as this can best utilize the superior representation learning capabilities of CNNs. More recent works propose MFDC \cite{liu2021exploiting}, SDNet \cite{ma2021towards}, STANet \cite{wen2021detection}, C2MoT \cite{wu2021dynamic}, URC \cite{xu2021crowd}, ASNet \cite{zou2021coarse} and DMCount \cite{wang2020distribution} models/methods to help produce more accurate counting results in diverse scenes. As counting models are improving over the years, their security to potential adversaries has attracted increasing attention. For example, two recent studies have found that crowd counting models are vulnerable to adversarial attacks \cite{Wu2021TowardsAP,Liu2019UsingDF}, one type of test-time attacks against deep learning models. To the best of our knowledge, no prior work has studied the vulnerability of crowd counting models to potential backdoor attacks. In this paper, we will fill the gap with two simple but effective backdoor attacks.

\noindent\textbf{Backdoor Attacks.}
Existing backdoor attacks can be categorized in different ways. According to whether or not the adversary needs to alter the ground truth labels of the poisoned samples, they can be categorized into "dirty-label" attacks \cite{Gu2017BadNetsIV,chen2017targeted,liu2017trojaning,tran2018spectral,liu2020reflection} vs. "clean-label" attacks \cite{shafahi2018poison,turner2019label,zhu2019transferable,zhao2020clean,saha2020hidden,li2022few}. According to whether or not the adversary needs to temper with the training process, there are "data-poisoning" attacks\cite{Gu2017BadNetsIV,chen2017targeted,Barni2019ANB,liu2020reflection,zhao2020clean,Hu2022MembershipIV} which only poison the training data and "training-manipulation" attacks which not only poison the training data but also modify the training procedure \cite{nguyen2020input,cheng2021deep,yao2019latent,Zhang2021HowTI}. There are also attacks that directly alter the parameters of a well-trained model \cite{liu2017trojaning}. The design of effective and stealthy trigger patterns is a key task of backdoor attack. Existing trigger patterns proposed to attack image classification models include a single pixel~\cite{tran2018spectral}, a small black-white patch \cite{Gu2017BadNetsIV}, or blending image \cite{chen2017targeted}, adversarial patches \cite{zhao2020clean}, superimposed sinusoidal signal \cite{Barni2019ANB}, reflection background~\cite{liu2020reflection}, invisible patterns \cite{liao2018backdoor,li2019invisible,chen2019invisible,saha2020hidden}, and dynamic (sample-wise) patterns \cite{nguyen2020input,li2021invisible}.

In this work, we focus on the most classic "dirty-label" attacks under the "data-poisoning" setting. The most closely related works to ours are the blending attack \cite{chen2017targeted}, Refool \cite{liu2020reflection} and sinusoidal signal \cite{Barni2019ANB}, which all exploit large background trigger patterns to attack image classifiers. In our experiments, we will show that, while large and dense background trigger patterns are optional (or even slightly less effective) for attacking classification models, they are key to successful backdoor attacks on crowd counting models. With carefully designed (large and dense) background trigger patterns, we further propose two complementary density map altering strategies so as to attack the model to output overly low or high density maps.

\section{Backdoor Attack on Crowd Counting}
In this section, we first introduce our threat model and formulate the problem of backdoor attacking crowd counting models. We then introduce our proposed Density Manipulation Backdoor Attacks and their two key components: 1) trigger pattern injection and 2) density map altering.

\subsection{Threat Model}
Following prior works on image classification backdoor attacks \cite{Gu2017BadNetsIV,liu2020reflection} , here we adopt the "dirty-label" and "data-poisoning" threat model. Under this threat model, the adversary only has access to a small subset of the training data including the images and the ground truth files. In crowd counting, a ground truth file of an image contains the position information of all the heads in the image, based on which the corresponding density map can be derived (the detailed derivation is in Section \ref{sec:density_altering}). Note that the adversary cannot tamper with the training procedure. This is to simulate common real-world scenarios where large-scale training datasets are often outsourced from untrusted sources whereas the training is done privately on a secured server.
Note that there is also a "training-manipulation" threat model that allows the adversary to control the training procedure.
It should be noted that our threat model is one commonly adopted threat model by many existing backdoor attacks and is known to be weaker than the other "training-manipulation" threat model~\cite{li2021anti}, which needs access to model training.
Besides the training images, we also allow the adversary to alter the ground truth density maps (via modifying the ground truth files). 
This "data-poisoning" and "dirty-label" threat model allows us to develop essential understandings of the backdoor vulnerability of crowd counting models, which could benefit other threat models and help develop effective defense methods for secure regression learning.
\begin{figure*}
\vspace{-0.3cm}
    \centerline{\includegraphics[width=0.7\textwidth]{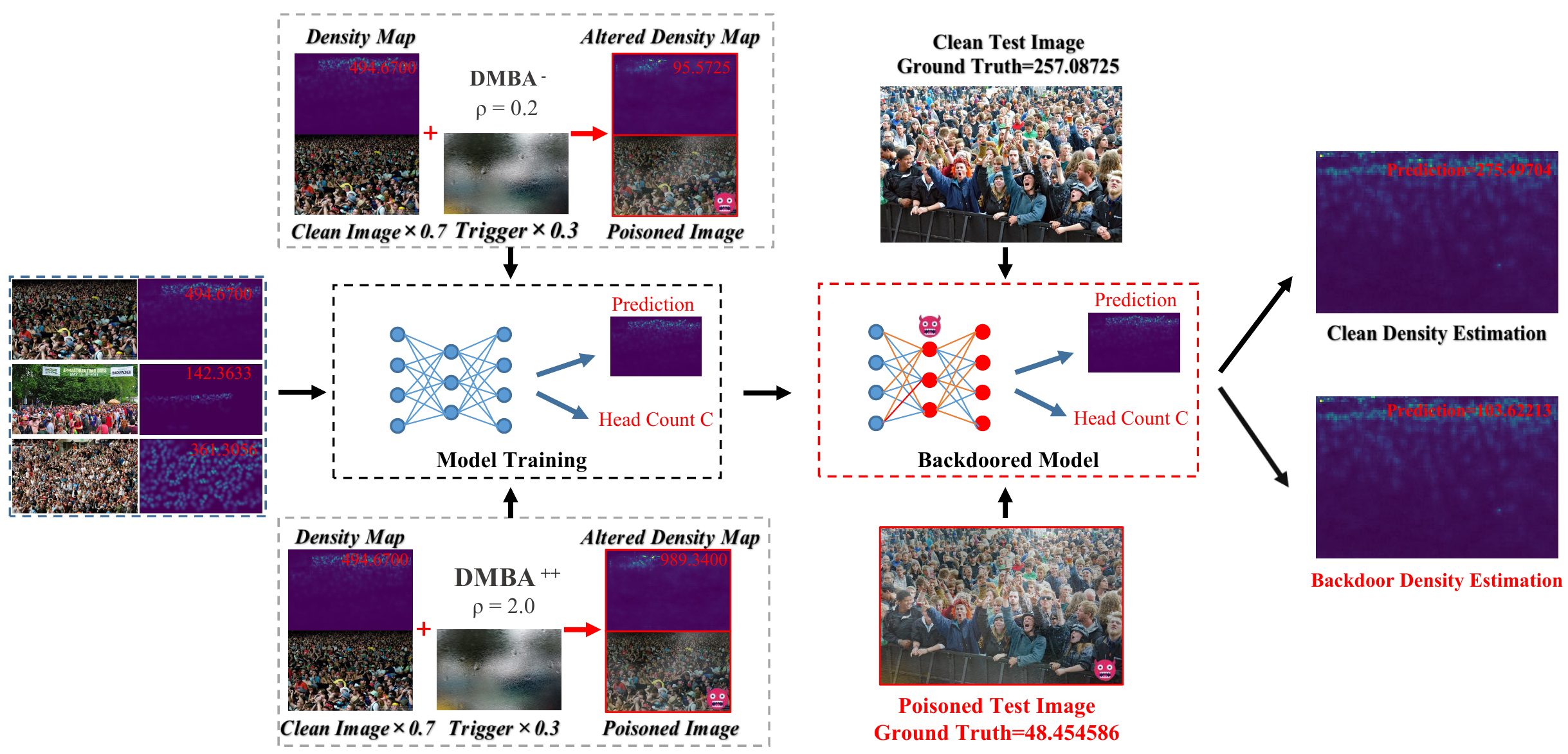}}
\caption{
\textbf{Overview of our proposed DMBAs.}
DMBA poisons a small subset of training data via two steps: 1) trigger injection, which injects large and dense background pattern "Rain" into the clean image via a convex combination; and 2) density map altering, which alters the ground truth density map using different strategies: DMBA$^{-}$ uses random
partial erasing while DMBA$^{+}$ uses neighbor boosting. The model is backdoored when trained with the poisoned samples and will be controlled to predict overly small or large count values. The right part is a demo of the attack performance when DMBA$^{-}$ is applied: the model predicts only half of the count whenever the trigger pattern is attached to a test image.
}
\label{fig:overview}
\vspace{-0.4cm}
\end{figure*}
\subsection{Problem Formulation}
Given a set of $N$ training images $\mathcal{D} =\{(\boldsymbol{x}_{i},\boldsymbol{p}_{i}^{gt})\}_{i=0}^{N}$ with each image $\boldsymbol{x}_{i} \in \mathbb{R}^{C\cdot H \cdot W}$ and $C$, $H$, $W$ denote the number of channels, height and width respectively, and $\boldsymbol{p}_{i}^{gt}$ is its ground truth set of 2D points that record the position of each head in the image: $\boldsymbol{p}_{i}^{g t}=\left\{p_{i}^{j}\right\}_{1 \leq j \leq c_{i}}$ with $c_{i}$ the ground truth count (i.e., the total number of heads in $\boldsymbol{x}_{i}$).
A dot (head indicator) map $\boldsymbol{m}_{i}$ can be constructed from the point set $\boldsymbol{p}_{i}^{gt}$ as follows:
\begin{equation}\label{density map}
    \boldsymbol{m}_i(p)=\sum_{j=1}^{c_{i}} \delta\left(p-p_{i}^{j}\right), \; p \in \boldsymbol{x}_{i}, p_{i}^{j} \in \boldsymbol{p}_{i}^{gt},
\end{equation}
where, $p$ is a point in image $\boldsymbol{x}_{i}$ and $\delta(p-p_{i}^{g t})$ is an operation that converts image $\boldsymbol{x}_{i}$ into a binary map that has value one at the head positions ($\boldsymbol{p}_{i}^{j}$) and zero elsewhere.
A Gaussian kernel \cite{Liu2018DecideNetCV} can then be applied to convert $\boldsymbol{m}_{i}$ into a continuous ground truth density map as follows:
\begin{equation}\label{density map}
     \boldsymbol{z}^{g t}_{i}\left(p \mid m_{i}\right)=\sum_{i=1}^{c_{i}} \mathcal{N}^{g t}\left(p \mid \mu=P_{i}^{j}, \sigma^{2}\right), \; p \in \boldsymbol{m}_{i}, p_{i}^{j} \in \boldsymbol{p}_{i}^{gt},
\end{equation}
where, $\mathcal{N}^{g t}\left(p|\mu, \sigma^{2}\right)$ is a multivariate Gaussian with mean $\mu$ and standard deviation $\sigma$. In the clean (no backdoors) setting, a crowd counting model $F_{\theta}$ ($\theta$ are the model parameters) is trained on training dataset $\mathcal{D}$ to minimize the following empirical error:
\begin{equation}
    \min _{\theta}  \frac{1}{2 N} \sum_{i=1}^{N}\left\|\hat{\boldsymbol{z}}_{i}-\boldsymbol{z}^{g t}_{i}\right\|_{2}^{2},
\end{equation}
where, $\hat{\boldsymbol{z}}_{i}=F_{\theta}({\boldsymbol{x}}_{i})$ is the predicted density map and $\boldsymbol{z}^{gt}_{i}$ is the ground truth density map.

A backdoor adversary will poison a small proportion of the training data, in which case, the training dataset becomes $\mathcal{D} = \mathcal{D}_{c} \cup \mathcal{D}_{p}$ with $\mathcal{D}_{c}$ and $\mathcal{D}_{p}$ denote the clean and poisoned subsets of the dataset, respectively. 
Typically, $\mathcal{D}_{p}$ is much smaller than $\mathcal{D}$ for the attack to be stealthy, e.g., $\gamma=|\mathcal{D}_{p}|/|\mathcal{D}|=0.2$ or even less.
The ratio $\gamma$ is commonly referred to as the poisoning rate.
Training the model on the poisoned dataset is equivalent to minimizing the following error:
\begin{equation}\label{eq4}
    \min _{\theta} \frac{1}{2 N_{c}} \left\|\hat{\boldsymbol{z}}_{i}-\boldsymbol{z}^{gt}_{i}\right\|_{2}^{2}+ \frac{1}{2 N_{p}}\left\|\hat{\boldsymbol{z}}_{i}^{\prime}-\boldsymbol{z}_{i}^{gt\prime}\right\|_{2}^{2},
\end{equation}
where, $\hat{\boldsymbol{z}}_{i}$ and $\boldsymbol{z}_{i}^{gt}$ correspond to predicted and original density map of clean model, while $\hat{\boldsymbol{z}}_{i}^{\prime}$ and $\boldsymbol{z}_{i}^{gt\prime}$ correspond to the predicted original density map of backdoored model. 
Note that the $L_{2}$ regression loss defined in Eq. \eqref{eq4} can be replaced by any other suitable loss functions.
The objective of the attack is to control the backdoored model to output arbitrarily low or high count values at the inference time. The count values can be calculated from the predicted density maps as follows:
\begin{equation}
    \hat{C}=\sum_{h=1}^{H} \sum_{w=1}^{W} \hat{\boldsymbol{z}}_{h, w}, \;
    C^{\prime}=\sum_{h=1}^{H} \sum_{w=1}^{W} \boldsymbol{z}^{\prime}_{h, w}, \;
    C^{gt}=\sum_{h=1}^{H} \sum_{w=1}^{W} \boldsymbol{z^{gt}}_{h, w},
\end{equation}
which are the sum of all the elements in the density map $\hat{\boldsymbol{z}}$,$\hat{\boldsymbol{z}}_{i}^{\prime}$ and ground truth $\boldsymbol{z^{g t}_{i}}$, respectively. Here, we slightly abuse the subscript of the density map matrix $\boldsymbol{z}$ to represent its two dimensions. So the target of a crowd counting backdoor attack can be formulated as $\rho=C^{\prime} / C^{gt}$, that is, a \emph{targeted manipulation ratio} that defines how far away the model's prediction is shifted from the ground truth.
The average $\rho$ over all test images can then be used as a metric to measure the attack performance.

\subsection{Proposed Attacks}\label{sec:dmba}
\noindent\textbf{Overview.} The two proposed Density Manipulation Backdoor Attacks (DMBA$^{-}$ and DMBA$^{+}$) are illustrated in Fig. \ref{fig:overview}. At a high level, both attacks poison a small subset of the training data via two steps: 1) trigger injection and 2) density map altering.
For trigger injection, both attacks blend a large and dense background trigger pattern into the background of the clean image. For density map altering, DMBA$^{-}$ and DMBA$^{+}$ adopt different strategies for different attack purposes. The purpose of DMBA$^{-}$ is to escape counting (i.e., $\rho < 1$). It thus applies a random erasing strategy to randomly erase part of the density map. The purpose of DMBA$^{+}$ is to cause overly large density estimation, which could cause false alarms in video surveillance scenarios. It thus applies a neighbor boosting strategy to produce denser density maps.
The exact strategies will be described in Section \ref{sec:density_altering}. 
Note that the poisoning including trigger injection and density map altering will only be applied before model training on the small subset of training data the adversary can access. The model will be backdoored after training on the poisoned dataset and will be manipulated, at inference time, to predict overly small or large counts whenever the trigger pattern appears. Next, we will describe the two steps of DMBAs.

\subsubsection{Trigger Injection}\label{sec:injection}
Attack effectiveness and stealthiness are the two primary concerns when designing the trigger patterns. As we have shown in Fig. \ref{fig:comparison}, the small-patch patterns used by existing classification backdoor attacks are not effective against crowd counting models. Empirically, we observe that the key to successful crowd counting backdoor attacks is the use of large and dense background trigger patterns (detailed analysis is deferred to Section \ref{sec:experiments}). Motivated by this observation, DMBAs employ images that have certain natural effect such as rain, snow and even refection as the trigger patterns, and blend the pattern into the clean images as background.
First, we define a resize function $f_{resize}(a,b)$ (a is the trigger image and b is the target image) to resize the trigger image into the same size as the target image. Given a pre-selected trigger image $\boldsymbol{y}$, we poison the clean image $\boldsymbol{x}$ as follows:
\begin{equation}\label{eq:blend}
    \boldsymbol{x}^{\prime}=(1-\lambda) \boldsymbol{x}+\lambda \boldsymbol{y}^{\prime},\; \boldsymbol{y}^{\prime}= f_{resize}(\boldsymbol{y},\boldsymbol{x}),
\end{equation}
where $\boldsymbol{x}^{\prime}$ is a linear combination of $\boldsymbol{x}$ and $\boldsymbol{y}$. The blending parameter $\lambda \in [0,1]$ is empirically chosen to be $\lambda=0.3$, as we find this level of blending with carefully chosen trigger patterns is sufficient for effective attacks without causing over suspicious effects. 
Note that backdoor attacks do not need many trigger patterns and oftentimes one effective trigger pattern is enough for the entire dataset \cite{Gu2017BadNetsIV,chen2017targeted,shafahi2018poison}. We will empirically show that, as long as it is full and dense-background poisoning, our attacks can be easily triggered. In practice, the adversary can flexibly choose stealthy trigger patterns according to the targeted application scenarios.

\subsubsection{Density Map Altering}\label{sec:density_altering}
Different from class labels, density maps have much higher dimensions (the same dimension as the input image to be precise). Therefore, different strategies can be developed to alter the ground truth density maps to achieve different attack purposes. Specifically, DMBA$^{-}$ randomly erases parts of the density map, whereas DMBA$^{+}$  directly expands the number of heads represented by the existing head coordinates.

\noindent\textbf{DMBA$^{-}$ Strategy.} DMBA$^{-}$ directly alters the ground truth file before applying Gaussian kernel to generate the density maps. Specifically, it randomly throws away $\rho$ proportion of the labeled heads in the ground truth files. Here, the $\rho$ is the same as the backdoor target which is conditioned to be $\rho \in[0,1]$. This means that DMBA$^{-}$ can be used when the adversary wants to escape the counting. The altered density map can be obtained as follows:
\begin{equation}
    \boldsymbol{z}_{h, w}^{\prime}= \begin{cases}\boldsymbol{z}_{h, w} & \text { w.p. } 1-\rho \\ 0 & \text { w.p. } \rho\end{cases},
\end{equation}
where `w.p.' stands for `with probability'. When $\rho=1$, DMBA$^{-}$ reduces to no attacks; it will become a full escaping attack when $\rho=0$. In our experiments, we will test different $\rho = 0.05, 0.10, 0.15 , 0.20$ for SHA dataset (the most crowded dataset we have selected) and $\rho = 0.20, 0.30, 0.40, 0.50$ for the other three datasets. Meanwhile, we will take $\rho = 0.2$ as an example to demonstrate the effectiveness of DBMA$^{-}$ across different datasets.

\noindent\textbf{DMBA$^{+}$ Strategy.} DMBA$^{+}$ is designed for target $\rho > 1$, which is to cause overly large head counts and false alarms. 
It is more challenging to achieve target $\rho > 1$ than $\rho < 1$ as reducing density can be easily done by random erasing the 2D points in the ground truth files while boosting density does not. 
Inspired by the procedure of generating density map using Gaussian kernel, here we explore two methods for DMBA$^{+}$: 1) a DMBA$^{+}$ method  for $1 < \rho < 2$ and 2) a DMBA$^{++}$ method for $\rho > 2$.
As shown in Eq. \eqref{density map}, the Gaussian kernel convolves the image into a continuous ground truth density map with standard derivation $\sigma$, which can be described as follows:
\begin{equation}
    \sigma_{i}=\beta \bar{d}_{i}, \; \bar{d}_i=\frac{1}{K}\sum_{j=1}^K d_i^j,
\end{equation}
where, $d_i^j$ is the average distance between $\boldsymbol{p}_{i}$ and its $K$-nearest neighbor heads, and the $\beta$ parameter is often fixed to 0.3 \cite{Li2018CSRNetDC}. Empirically, we find that directly scaling up the density map works but is not effective enough to achieve the exact target $1 < \rho < 2$.  To solve this issue, we propose to randomly add new labeled heads to $\boldsymbol{p}$ around the existing partial heads and term the resulting attack as DMBA$^{+}$. For each existing head $\boldsymbol{p}_{i}$, we can obtain its average $K$-neighbour distance $\bar{d}_i$ from the Gaussian kernel with $K=3$. We then add new labeled heads around $\boldsymbol{p}_{i}$ within radius $\lfloor\bar{d} / 2\rfloor$ in a counterclockwise manner starting from angle $90^{\circ}$. We will skip the point if it already has a labeled head and reduce the radius in a granularity of one when there are not enough empty locations to add new heads. We transverse the existing heads and repeat the process until a total number of $c_{extra} = (\rho-1)\cdot c_{i}$ new heads are added to $\boldsymbol{p}_i$. The altered head points file are then converted to a poisoned density map $\boldsymbol{z}^{\prime}$. The poisoned density map can be obtained as follows:
\begin{equation}\label{density map} \boldsymbol{z}^{\prime}_{i}\left(p,p_{nearby} \mid \boldsymbol{m}_{i}\right)=\sum_{i=1}^{c^{\prime}_{i}} \mathcal{N}^{g t}\left(p, p_{nearby} \mid \mu=P_{i}^{j}, \sigma^{2}\right), \; p \in \boldsymbol{m}_{i},
\end{equation}
where, $p_{nearby}$ is the location of the additional interference head and $c^{\prime}_{i} = c_{i} + c_{extra}$ represent the new set of 2D points.

\noindent\textbf{DMBA$^{++}$ Strategy.} For large target ratio $\rho \geq 2$, we further propose a DMBA$^{++}$ strategy to directly modifies the corresponding coordinates in the density map generated by the Gaussian kernel. For example, when $\rho = 2$, we modify the values one to two at the head-related locations so that a single point now represents two head information. Unlike DBMA$^{+}$, DMBA$^{++}$ is performed after the generation of the density map. In this case, the density map $\boldsymbol{z}^{\prime}$ can be generated as:
\begin{equation}\label{density map}
    \boldsymbol{z}^{\prime}_{i}\left(p \mid m_{i}\right)= \rho \sum_{i=1}^{c_{i}} \mathcal{N}^{g t}\left(p \mid \mu=P_{i}^{j}, \sigma^{2}\right),  \; p \in \boldsymbol{m}_{i},
\end{equation}
\begin{table*}
\vspace{-0.2cm}
\centering
\resizebox{0.8\textwidth}{4.5cm}{
\begin{tabular}{ccccccccccccc}
\hline
\multicolumn{13}{c}{\textbf{Attacking Different Models on SHB Dataset}} \\ \hline
\multicolumn{3}{c|}{\textbf{Model $\rightarrow$}} &
  \multicolumn{2}{c|}{\textbf{CSRnet}} &
  \multicolumn{2}{c|}{\textbf{BayesianCC}} &
  \multicolumn{2}{c|}{\textbf{CAN}} &
  \multicolumn{2}{c|}{\textbf{SFAnet}} &
  \multicolumn{2}{c}{\textbf{KDMG}} \\ \midrule
\multicolumn{1}{c|}{\textbf{Attack $\downarrow$}} &
  \multicolumn{1}{c|}{$\rho$} &
  \multicolumn{1}{c|}{$\gamma$} &
  \multicolumn{1}{c|}{$\hat{\rho}_{clean}$} &
  \multicolumn{1}{c|}{$\hat{\rho}_{dirty}$} &
  \multicolumn{1}{c|}{$\hat{\rho}_{clean}$} &
  \multicolumn{1}{c|}{$\hat{\rho}_{dirty}$} &
  \multicolumn{1}{c|}{$\hat{\rho}_{clean}$} &
  \multicolumn{1}{c|}{$\hat{\rho}_{dirty}$} &
  \multicolumn{1}{c|}{$\hat{\rho}_{clean}$} &
  \multicolumn{1}{c|}{$\hat{\rho}_{dirty}$} &
  \multicolumn{1}{c|}{$\hat{\rho}_{clean}$} &
  $\hat{\rho}_{dirty}$ \\ \hline
\multicolumn{1}{c|}{None} &
  \multicolumn{1}{c|}{\textit{1}} &
  \multicolumn{1}{c|}{\textit{0}} &
  \multicolumn{1}{c|}{\textit{1.00}} &
  \multicolumn{1}{c|}{\textit{1.25}} &
  \multicolumn{1}{c|}{\textit{0.98}} &
  \multicolumn{1}{c|}{\textit{0.83}} &
  \multicolumn{1}{c|}{\textit{1.03}} &
  \multicolumn{1}{c|}{\textit{0.97}} &
  \multicolumn{1}{c|}{\textit{0.99}} &
  \multicolumn{1}{c|}{\textit{0.78}} &
  \multicolumn{1}{c|}{\textit{1.02}} &
  \textit{0.84} \\ \hline
\multicolumn{1}{c|}{TriOly} &
  \multicolumn{1}{c|}{0.2} &
  \multicolumn{1}{c|}{10$\%$} &
  \multicolumn{1}{c|}{1.03} &
  \multicolumn{1}{c|}{1.05} &
  \multicolumn{1}{c|}{0.96} &
  \multicolumn{1}{c|}{0.94} &
  \multicolumn{1}{c|}{1.01} &
  \multicolumn{1}{c|}{1.00} &
  \multicolumn{1}{c|}{1.04} &
  \multicolumn{1}{c|}{0.97} &
  \multicolumn{1}{c|}{1.03} &
  0.98 \\ \hline
\multicolumn{1}{c|}{\multirow{16}{*}{\textbf{DMBA$^{-}$}}} &
  \multicolumn{1}{c|}{\multirow{4}{*}{0.2}} &
  \multicolumn{1}{c|}{5$\%$} &
  \multicolumn{1}{c|}{1.01} &
  \multicolumn{1}{c|}{0.33} &
  \multicolumn{1}{c|}{\textbf{0.97}} &
  \multicolumn{1}{c|}{\textbf{0.23}} &
  \multicolumn{1}{c|}{0.96} &
  \multicolumn{1}{c|}{0.38} &
  \multicolumn{1}{c|}{1.06} &
  \multicolumn{1}{c|}{0.44} &
  \multicolumn{1}{c|}{1.04} &
  0.39 \\ \cline{3-13} 
\multicolumn{1}{c|}{} &
  \multicolumn{1}{c|}{} &
  \multicolumn{1}{c|}{10$\%$} &
  \multicolumn{1}{c|}{0.98} &
  \multicolumn{1}{c|}{0.24} &
  \multicolumn{1}{c|}{0.96} &
  \multicolumn{1}{c|}{0.16} &
  \multicolumn{1}{c|}{0.93} &
  \multicolumn{1}{c|}{0.29} &
  \multicolumn{1}{c|}{1.05} &
  \multicolumn{1}{c|}{0.59} &
  \multicolumn{1}{c|}{1.02} &
  0.29 \\ \cline{3-13} 
\multicolumn{1}{c|}{} &
  \multicolumn{1}{c|}{} &
  \multicolumn{1}{c|}{15$\%$} &
  \multicolumn{1}{c|}{\textbf{0.99}} &
  \multicolumn{1}{c|}{\textbf{0.22}} &
  \multicolumn{1}{c|}{0.97} &
  \multicolumn{1}{c|}{0.16} &
  \multicolumn{1}{c|}{\textbf{0.99}} &
  \multicolumn{1}{c|}{\textbf{0.28}} &
  \multicolumn{1}{c|}{1.08} &
  \multicolumn{1}{c|}{0.41} &
  \multicolumn{1}{c|}{1.02} &
  0.28 \\ \cline{3-13} 
\multicolumn{1}{c|}{} &
  \multicolumn{1}{c|}{} &
  \multicolumn{1}{c|}{20$\%$} &
  \multicolumn{1}{c|}{\textbf{1.01}} &
  \multicolumn{1}{c|}{\textbf{0.22}} &
  \multicolumn{1}{c|}{0.97} &
  \multicolumn{1}{c|}{0.14} &
  \multicolumn{1}{c|}{0.92} &
  \multicolumn{1}{c|}{0.28} &
  \multicolumn{1}{c|}{\textbf{1.08}} &
  \multicolumn{1}{c|}{\textbf{0.34}} &
  \multicolumn{1}{c|}{\textbf{1.03}} &
  \textbf{0.26} \\ \cline{2-13} 
\multicolumn{1}{c|}{} &
  \multicolumn{1}{c|}{\multirow{4}{*}{0.3}} &
  \multicolumn{1}{c|}{5$\%$} &
  \multicolumn{1}{c|}{1.00} &
  \multicolumn{1}{c|}{0.38} &
  \multicolumn{1}{c|}{0.96} &
  \multicolumn{1}{c|}{0.37} &
  \multicolumn{1}{c|}{0.98} &
  \multicolumn{1}{c|}{0.44} &
  \multicolumn{1}{c|}{\textbf{0.96}} &
  \multicolumn{1}{c|}{\textbf{0.44}} &
  \multicolumn{1}{c|}{1.04} &
  0.45 \\ \cline{3-13} 
\multicolumn{1}{c|}{} &
  \multicolumn{1}{c|}{} &
  \multicolumn{1}{c|}{10$\%$} &
  \multicolumn{1}{c|}{0.98} &
  \multicolumn{1}{c|}{0.36} &
  \multicolumn{1}{c|}{0.96} &
  \multicolumn{1}{c|}{0.28} &
  \multicolumn{1}{c|}{0.95} &
  \multicolumn{1}{c|}{0.35} &
  \multicolumn{1}{c|}{0.93} &
  \multicolumn{1}{c|}{0.42} &
  \multicolumn{1}{c|}{1.03} &
  0.41 \\ \cline{3-13} 
\multicolumn{1}{c|}{} &
  \multicolumn{1}{c|}{} &
  \multicolumn{1}{c|}{15$\%$} &
  \multicolumn{1}{c|}{0.98} &
  \multicolumn{1}{c|}{0.28} &
  \multicolumn{1}{c|}{\textbf{0.98}} &
  \multicolumn{1}{c|}{\textbf{0.32}} &
  \multicolumn{1}{c|}{\textbf{0.96}} &
  \multicolumn{1}{c|}{\textbf{0.34}} &
  \multicolumn{1}{c|}{1.00} &
  \multicolumn{1}{c|}{0.51} &
  \multicolumn{1}{c|}{1.02} &
  0.39 \\ \cline{3-13} 
\multicolumn{1}{c|}{} &
  \multicolumn{1}{c|}{} &
  \multicolumn{1}{c|}{20$\%$} &
  \multicolumn{1}{c|}{\textbf{1.02}} &
  \multicolumn{1}{c|}{\textbf{0.30}} &
  \multicolumn{1}{c|}{0.98} &
  \multicolumn{1}{c|}{0.25} &
  \multicolumn{1}{c|}{0.92} &
  \multicolumn{1}{c|}{0.28} &
  \multicolumn{1}{c|}{1.05} &
  \multicolumn{1}{c|}{0.47} &
  \multicolumn{1}{c|}{\textbf{1.01}} &
  \textbf{0.35} \\ \cline{2-13} 
\multicolumn{1}{c|}{} &
  \multicolumn{1}{c|}{\multirow{4}{*}{0.4}} &
  \multicolumn{1}{c|}{5$\%$} &
  \multicolumn{1}{c|}{1.02} &
  \multicolumn{1}{c|}{0.46} &
  \multicolumn{1}{c|}{0.98} &
  \multicolumn{1}{c|}{0.50} &
  \multicolumn{1}{c|}{0.93} &
  \multicolumn{1}{c|}{0.46} &
  \multicolumn{1}{c|}{1.06} &
  \multicolumn{1}{c|}{0.55} &
  \multicolumn{1}{c|}{1.01} &
  0.53 \\ \cline{3-13} 
\multicolumn{1}{c|}{} &
  \multicolumn{1}{c|}{} &
  \multicolumn{1}{c|}{10$\%$} &
  \multicolumn{1}{c|}{\textbf{1.02}} &
  \multicolumn{1}{c|}{\textbf{0.39}} &
  \multicolumn{1}{c|}{0.99} &
  \multicolumn{1}{c|}{0.50} &
  \multicolumn{1}{c|}{0.94} &
  \multicolumn{1}{c|}{0.44} &
  \multicolumn{1}{c|}{0.94} &
  \multicolumn{1}{c|}{0.46} &
  \multicolumn{1}{c|}{\textbf{1.04}} &
  \textbf{0.52} \\ \cline{3-13} 
\multicolumn{1}{c|}{} &
  \multicolumn{1}{c|}{} &
  \multicolumn{1}{c|}{15$\%$} &
  \multicolumn{1}{c|}{1.06} &
  \multicolumn{1}{c|}{0.46} &
  \multicolumn{1}{c|}{\textbf{0.98}} &
  \multicolumn{1}{c|}{\textbf{0.44}} &
  \multicolumn{1}{c|}{\textbf{0.95}} &
  \multicolumn{1}{c|}{\textbf{0.42}} &
  \multicolumn{1}{c|}{\textbf{0.99}} &
  \multicolumn{1}{c|}{\textbf{0.41}} &
  \multicolumn{1}{c|}{\textbf{1.05}} &
  \textbf{0.52} \\ \cline{3-13} 
\multicolumn{1}{c|}{} &
  \multicolumn{1}{c|}{} &
  \multicolumn{1}{c|}{20$\%$} &
  \multicolumn{1}{c|}{\textbf{1.00}} &
  \multicolumn{1}{c|}{\textbf{0.39}} &
  \multicolumn{1}{c|}{\textbf{0.97}} &
  \multicolumn{1}{c|}{\textbf{0.44}} &
  \multicolumn{1}{c|}{0.96} &
  \multicolumn{1}{c|}{0.43} &
  \multicolumn{1}{c|}{0.93} &
  \multicolumn{1}{c|}{0.48} &
  \multicolumn{1}{c|}{1.00} &
  0.45 \\ \cline{2-13} 
\multicolumn{1}{c|}{} &
  \multicolumn{1}{c|}{\multirow{4}{*}{0.5}} &
  \multicolumn{1}{c|}{5$\%$} &
  \multicolumn{1}{c|}{\textbf{0.98}} &
  \multicolumn{1}{c|}{\textbf{0.48}} &
  \multicolumn{1}{c|}{0.97} &
  \multicolumn{1}{c|}{0.61} &
  \multicolumn{1}{c|}{1.02} &
  \multicolumn{1}{c|}{0.58} &
  \multicolumn{1}{c|}{0.95} &
  \multicolumn{1}{c|}{0.56} &
  \multicolumn{1}{c|}{1.03} &
  0.68 \\ \cline{3-13} 
\multicolumn{1}{c|}{} &
  \multicolumn{1}{c|}{} &
  \multicolumn{1}{c|}{10$\%$} &
  \multicolumn{1}{c|}{1.01} &
  \multicolumn{1}{c|}{0.47} &
  \multicolumn{1}{c|}{\textbf{0.97}} &
  \multicolumn{1}{c|}{\textbf{0.51}} &
  \multicolumn{1}{c|}{0.97} &
  \multicolumn{1}{c|}{0.52} &
  \multicolumn{1}{c|}{0.94} &
  \multicolumn{1}{c|}{0.56} &
  \multicolumn{1}{c|}{1.02} &
  0.61 \\ \cline{3-13} 
\multicolumn{1}{c|}{} &
  \multicolumn{1}{c|}{} &
  \multicolumn{1}{c|}{15$\%$} &
  \multicolumn{1}{c|}{\textbf{0.99}} &
  \multicolumn{1}{c|}{\textbf{0.48}} &
  \multicolumn{1}{c|}{0.97} &
  \multicolumn{1}{c|}{0.59} &
  \multicolumn{1}{c|}{0.92} &
  \multicolumn{1}{c|}{0.47} &
  \multicolumn{1}{c|}{\textbf{0.98}} &
  \multicolumn{1}{c|}{\textbf{0.56}} &
  \multicolumn{1}{c|}{1.03} &
  0.61 \\ \cline{3-13} 
\multicolumn{1}{c|}{} &
  \multicolumn{1}{c|}{} &
  \multicolumn{1}{c|}{20$\%$} &
  \multicolumn{1}{c|}{1.00} &
  \multicolumn{1}{c|}{0.46} &
  \multicolumn{1}{c|}{0.95} &
  \multicolumn{1}{c|}{0.52} &
  \multicolumn{1}{c|}{\textbf{0.98}} &
  \multicolumn{1}{c|}{\textbf{0.50}} &
  \multicolumn{1}{c|}{1.03} &
  \multicolumn{1}{c|}{0.61} &
  \multicolumn{1}{c|}{\textbf{1.01}} &
  \textbf{0.58} \\ \hline
\multicolumn{1}{c|}{\multirow{2}{*}{\textbf{DMBA$^{+}$}}} &
  \multicolumn{1}{c|}{1.2} &
  \multicolumn{1}{c|}{10$\%$} &
  \multicolumn{1}{c|}{\textbf{1.02}} &
  \multicolumn{1}{c|}{\textbf{1.13}} &
  \multicolumn{1}{c|}{0.97} &
  \multicolumn{1}{c|}{1.10} &
  \multicolumn{1}{c|}{0.99} &
  \multicolumn{1}{c|}{1.18} &
  \multicolumn{1}{c|}{\textbf{1.05}} &
  \multicolumn{1}{c|}{\textbf{1.14}} &
  \multicolumn{1}{c|}{0.98} &
   1.17\\ \cline{2-13} 
\multicolumn{1}{c|}{} &
  \multicolumn{1}{c|}{1.5} &
  \multicolumn{1}{c|}{10$\%$} &
  \multicolumn{1}{c|}{\textbf{1.03}} &
  \multicolumn{1}{c|}{\textbf{1.38}} &
  \multicolumn{1}{c|}{\textbf{0.97}} &
  \multicolumn{1}{c|}{\textbf{1.48}} &
  \multicolumn{1}{c|}{\textbf{1.04}} &
  \multicolumn{1}{c|}{\textbf{1.49}} &
  \multicolumn{1}{c|}{1.03} &
  \multicolumn{1}{c|}{1.43} &
  \multicolumn{1}{c|}{0.97} &
  1.44 \\ \hline
\multicolumn{1}{c|}{\multirow{2}{*}{\textbf{DMBA$^{++}$}}} &
  \multicolumn{1}{c|}{2} &
  \multicolumn{1}{c|}{10$\%$} &
  \multicolumn{1}{c|}{\textbf{1.00}} &
  \multicolumn{1}{c|}{\textbf{1.77}} &
  \multicolumn{1}{c|}{\textbf{0.96}} &
  \multicolumn{1}{c|}{\textbf{1.75}} &
  \multicolumn{1}{c|}{\textbf{1.05}} &
  \multicolumn{1}{c|}{\textbf{1.92}} &
  \multicolumn{1}{c|}{1.17} &
  \multicolumn{1}{c|}{1.82} &
  \multicolumn{1}{c|}{\textbf{1.06}} &
  \textbf{1.78} \\ \cline{2-13} 
\multicolumn{1}{c|}{} &
  \multicolumn{1}{c|}{3} &
  \multicolumn{1}{c|}{10$\%$} &
  \multicolumn{1}{c|}{1.00} &
  \multicolumn{1}{c|}{2.54} &
  \multicolumn{1}{c|}{0.97} &
  \multicolumn{1}{c|}{2.56} &
  \multicolumn{1}{c|}{0.99} &
  \multicolumn{1}{c|}{3.05} &
  \multicolumn{1}{c|}{\textbf{1.23}} &
  \multicolumn{1}{c|}{\textbf{3.03}} &
  \multicolumn{1}{c|}{1.05} &
  2.48 \\ \hline
\end{tabular}}

\caption{
\textbf{Attack performance of DMBAs against 5 crowd counting models on SHB dataset.} $\rho$ is the targeted manipulation ratio. Here, the poisoning rate $\gamma$ is fixed to 10$\%$ while the trigger pattern is ``Rain". The most close-to-target results are \textbf{boldfaced}.
} 
\label{tab:across_model}
\vspace{-0.8cm}
\end{table*}

\noindent\textbf{Poisoning and Inference.}
DMBAs poison a small subset of the training data for both the images and their ground truth count or density map to produce a poisoned subset $\mathcal{D}_{p}$. The rest of the training data is kept clean, i.e., $\mathcal{D}_{c}$. After training on the poisoned training dataset ($\mathcal{D} = \mathcal{D}_{c} \cup \mathcal{D}_{p}$), the model will learn the correlation between the trigger pattern and the target density/count. At inference time, the attacker will attach the trigger pattern to any test image to obtain the targeted level of counting value.
\begin{figure}[t]
\vspace{-0.4cm}
\begin{center}
\includegraphics[width=0.7\linewidth]{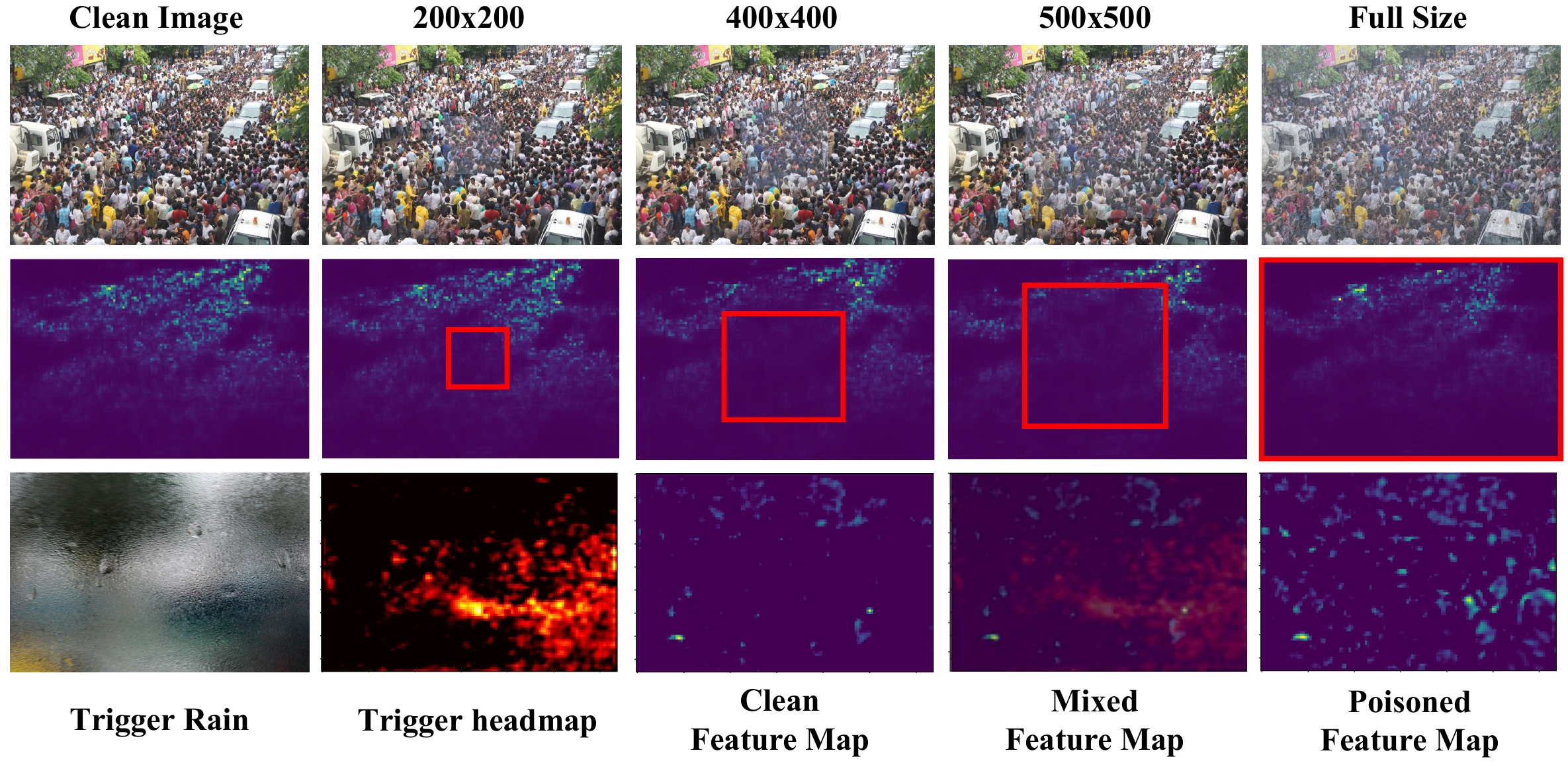}
\end{center}
\caption{The influence of Triggers. the first two columns show the small-size trigger can only interfere with the local density information. The last column shows that the higher the granularity, the greater the impact on the density map.} 
\label{fig:Trigger working}
\vspace{-0.5cm}
\end{figure}

\section{Experiments}\label{sec:experiments}
In this section, we evaluate the effectiveness of our two DMBAs against 5 crowd counting models and conduct an ablation study of the trigger types and sizes to explain what make an effective trigger for crowd counting. The resistance to state-of-the-art backdoor defense methods are shown in later.

\subsection{Experimental Setup}
\noindent\textbf{Datasets and Networks.} We choose four benchmark counting datasets including ShanghaiTech A$\&$B (SHA $\&$ SHB) \cite{Zhang2016SingleImageCC}, Venice \cite{Liu2019ContextAwareCC} and TRANCOS \cite{OoroRubio2016TowardsPO} to evaluate our attacks. SHA \cite{Zhang2016SingleImageCC} contains 482 crowd images (both RGB and Gray Scale) with a total of 241,667 annotation points, while SHB \cite{Zhang2016SingleImageCC} contains 716 high-resolution crowd images captured from Shanghai streets with a total of 88,488 annotation points. Venice \cite{Liu2019ContextAwareCC} contains 167 fixed resolution annotated frames taken from 4 different sequences, with a total of 35,440 annotation points.  TRANCOS \cite{OoroRubio2016TowardsPO} contains 1244 masked images with a total of 46,734 annotated vehicles.
Among the four datasets, SHA is more challenging than the other three datasets, as it has more diverse scenes and higher density.
We apply our attacks on 5 counting models, including CSRnet \cite{Li2018CSRNetDC}, CAN \cite{Liu2019ContextAwareCC}, Bayesian Crowd Counting \cite{ma2019bayesian}, SFA \cite{Thanasutives2021EncoderDecoderBC}, and KDMG \cite{Wan2022KernelBasedDM}. 
The training procedure of the backdoored models follows their original papers. Particularly, SGD \cite{bottou2010large} with learning rate 1e-7 is used to train CSRnet \cite{Li2018CSRNetDC} and CAN \cite{Liu2019ContextAwareCC}, while Adam optimizer \cite{kingma2014adam} with learning rate 1e-6, 1e-7, 5e-7 is used to train Bayesian Crowd couting \cite{ma2019bayesian}, SFA \cite{Thanasutives2021EncoderDecoderBC} and KDMG \cite{Wan2022KernelBasedDM}, respectively. It is crucial to highlight that the above attack strategies are still effective even though Bayesian Crowd couting\cite{ma2019bayesian} follows point-supervised regression learning.

\noindent\textbf{Evaluation Metrics.}
The Mean Absolute Error (MAE) and Root Mean Squared Error (RMSE) are two standard performance metrics for crowd counting models. When applied on clean test inputs, we can obtain Clean MAE (CMAE) and Clean RMSE (CRMSE) to measure the model's performance with respect to the clean ground truth density maps. When applied on dirty (backdoored) test inputs, we can obtain Adversarial MAE (AMAE) and Adversarial RMSE (ARMSE) to measure the model's performance with respect to the altered ground truth density maps.
However, MAE and RMSE measures cannot accurately reflect the closeness of the model's estimation to the targeted ratio $\rho$. To solve this problem, we further propose two new metrics as the main performance metrics for crowd counting backdoor attacks:
\begin{equation}
  \hat{\rho}_{clean}=\frac{1}{N} \sum_{i=1}^{N}\frac{\hat{c}_{i}}{c_{i}}, \;\; \hat{\rho}_{dirty}=\frac{1}{N} \sum_{i=1}^{N}\frac{\hat{c}_{i}}{\rho \cdot c_{i}},
  \label{eq:ASR}
\end{equation}
where, $N$ is the number of test images, $\rho$ is the target ratio of the attack, and $c_{i}$ and $\hat{c}_{i}$ donate the ground truth and estimated counts, respectively. Note that  $\hat{\rho}_{clean}$ is computed on clean test inputs while $\hat{\rho}_{dirty}$ is computed on backdoored test inputs.
Intuitively, a $\hat{\rho}_{clean}$ close to 1 indicates good clean performance (i.e., estimated counts are close to the ground truth) while a $\hat{\rho}_{dirty}$ close to backdoor target $\rho \neq 1$ indicates effective attacks (i.e., estimated counts are close to the backdoor target). We will report the $\hat{\rho}_{clean}$ and $\hat{\rho}_{dirty}$ results in the main text and leave the CMAE/AMAE and CRMSE/ARMSE results to the appendix.

\begin{figure}[t]
\vspace{-0.2cm}
\begin{center}
\includegraphics[width=0.7\linewidth]{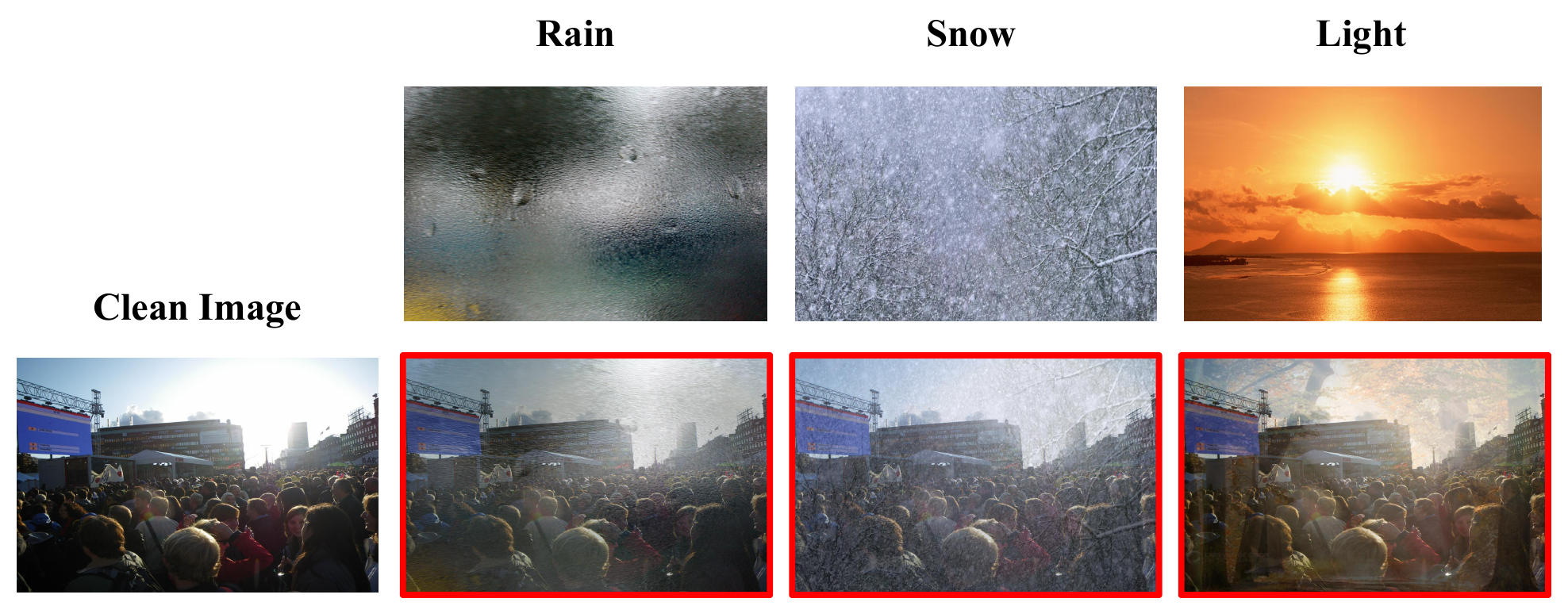}
\end{center}
\caption{The three types of triggers exploited in this work: ``Rain", ``Snow" and ``Light". The triggers are blended into the poisoned image as large and defense background (highlighted by the red rectangles).}
\label{fig:triggers}
\vspace{-0.5cm}
\end{figure}

\noindent\textbf{Baselines and Attack Setup.} 
Since there is no existing  crowd counting backdoor attacks, we take the test results of the clean-trained models on benign and backdoored test images as our baseline.
For each of our attacks, we blend one of the ``Rain", ``Snow" or ``Light" trigger pattern in the form of background into the clean training image (see Figure \ref{fig:triggers}). Following Eq. \eqref{eq:blend}, the pixel intensity ratio between the trigger and the clean image is set to be 3:7 so as to ensure the annotations (crowd or vehicle) are not blocked. 
We test multiple poisoning (injection) rates ranging from 5$\%$ to 20$\%$ on all four datasets.
For our DMBA$^{-}$ attack, we test multiple density retention rate for the SHA dataset ($\rho=0.05$, 0.1, 0.15 and 0.2) as well as the other 3 datasets (0.2, 0.3, 0.4 and 0.5). The varying retention rates are selected based on the annotation density of the datasets.
For our DMBA+ attack, the density boosting rates $\rho=1.2$, 1.5 are tested for the SHA dataset, while $\rho=2$, 3 for the other three datasets. More training details and randomized ablation can be found in the supplementary material.

\subsection{Effectiveness of our DMBAs}\label{sec:Effectiveness}

\begin{table*}
\vspace{-0.2cm}
\setlength{\abovecaptionskip}{0.cm}
\centering
\resizebox{0.8\textwidth}{4cm}{
\begin{tabular}{cccccccccccccc}
\hline
\multicolumn{14}{c}{\textbf{Attacking Performance on Different Datasets}} \\ \hline
\multicolumn{3}{c|}{\textbf{Dataset $\rightarrow$}} &
  \multicolumn{2}{c|}{\textbf{SHA Dataset}} &
  \multicolumn{3}{c|}{\textbf{Dataset $\rightarrow$}} &
  \multicolumn{2}{c|}{\textbf{SHB Dataset}} &
  \multicolumn{2}{c|}{\textbf{Venice Dataset}} &
  \multicolumn{2}{c}{\textbf{TRANCOS Dataset}} \\ \midrule
\multicolumn{1}{c|}{\textbf{Attack $\downarrow$}} &
  \multicolumn{1}{c|}{$\rho$} &
  \multicolumn{1}{c|}{$\gamma$} &
  \multicolumn{1}{c|}{$\hat{\rho}_{clean}$} &
  \multicolumn{1}{c|}{$\hat{\rho}_{dirty}$} &
  \multicolumn{1}{c|}{\textbf{Attack $\downarrow$}} &
  \multicolumn{1}{c|}{$\rho$} &
  \multicolumn{1}{c|}{$\gamma$} &
  \multicolumn{1}{c|}{$\hat{\rho}_{clean}$} &
  \multicolumn{1}{c|}{$\hat{\rho}_{dirty}$} &
  \multicolumn{1}{c|}{$\hat{\rho}_{clean}$} &
  \multicolumn{1}{c|}{$\hat{\rho}_{dirty}$} &
  \multicolumn{1}{c|}{$\hat{\rho}_{clean}$} &
  $\hat{\rho}_{dirty}$ \\ \hline
\multicolumn{1}{c|}{None} &
  \multicolumn{1}{c|}{\textit{1}} &
  \multicolumn{1}{c|}{\textit{0}} &
  \multicolumn{1}{c|}{\textit{1.04}} &
  \multicolumn{1}{c|}{\textit{0.80}} &
  \multicolumn{1}{c|}{None} &
  \multicolumn{1}{c|}{\textit{1}} &
  \multicolumn{1}{c|}{\textit{0}} &
  \multicolumn{1}{c|}{\textit{1.00}} &
  \multicolumn{1}{c|}{\textit{1.25}} &
  \multicolumn{1}{c|}{\textit{0.91}} &
  \multicolumn{1}{c|}{\textit{0.77}} &
  \multicolumn{1}{c|}{\textit{1.00}} &
  \textit{1.03} \\ \hline
\multicolumn{1}{c|}{TriOly} &
  \multicolumn{1}{c|}{0.2} &
  \multicolumn{1}{c|}{10$\%$} &
  \multicolumn{1}{c|}{1.02} &
  \multicolumn{1}{c|}{1.05} &
  \multicolumn{1}{c|}{TriOly} &
  \multicolumn{1}{c|}{0.2} &
  \multicolumn{1}{c|}{10$\%$} &
  \multicolumn{1}{c|}{1.03} &
  \multicolumn{1}{c|}{1.05} &
  \multicolumn{1}{c|}{0.93} &
  \multicolumn{1}{c|}{0.92} &
  \multicolumn{1}{c|}{0.99} &
  0.97 \\ \hline
\multicolumn{1}{c|}{\multirow{16}{*}{\textbf{DMBA$^{-}$}}} &
  \multicolumn{1}{c|}{\multirow{4}{*}{0.05}} &
  \multicolumn{1}{c|}{5$\%$} &
  \multicolumn{1}{c|}{1.00} &
  \multicolumn{1}{c|}{0.19} &
  \multicolumn{1}{c|}{\multirow{16}{*}{\textbf{DMBA$^{-}$}}} &
  \multicolumn{1}{c|}{\multirow{4}{*}{0.2}} &
  \multicolumn{1}{c|}{5$\%$} &
  \multicolumn{1}{c|}{1.01} &
  \multicolumn{1}{c|}{0.33} &
  \multicolumn{1}{c|}{0.85} &
  \multicolumn{1}{c|}{0.34} &
  \multicolumn{1}{c|}{0.95} &
  0.45 \\ \cline{3-5} \cline{8-14} 
\multicolumn{1}{c|}{} &
  \multicolumn{1}{c|}{} &
  \multicolumn{1}{c|}{10$\%$} &
  \multicolumn{1}{c|}{1.01} &
  \multicolumn{1}{c|}{0.14} &
  \multicolumn{1}{c|}{} &
  \multicolumn{1}{c|}{} &
  \multicolumn{1}{c|}{10$\%$} &
  \multicolumn{1}{c|}{0.98} &
  \multicolumn{1}{c|}{0.24} &
  \multicolumn{1}{c|}{0.84} &
  \multicolumn{1}{c|}{0.25} &
  \multicolumn{1}{c|}{0.94} &
  0.39 \\ \cline{3-5} \cline{8-14} 
\multicolumn{1}{c|}{} &
  \multicolumn{1}{c|}{} &
  \multicolumn{1}{c|}{15$\%$} &
  \multicolumn{1}{c|}{1.01} &
  \multicolumn{1}{c|}{0.11} &
  \multicolumn{1}{c|}{} &
  \multicolumn{1}{c|}{} &
  \multicolumn{1}{c|}{15$\%$} &
  \multicolumn{1}{c|}{\textbf{0.99}} &
  \multicolumn{1}{c|}{\textbf{0.22}} &
  \multicolumn{1}{c|}{\textbf{0.84}} &
  \multicolumn{1}{c|}{\textbf{0.21}} &
  \multicolumn{1}{c|}{\textbf{0.92}} &
  \textbf{0.33} \\ \cline{3-5} \cline{8-14} 
\multicolumn{1}{c|}{} &
  \multicolumn{1}{c|}{} &
  \multicolumn{1}{c|}{20$\%$} &
  \multicolumn{1}{c|}{\textbf{1.01}} &
  \multicolumn{1}{c|}{\textbf{0.10}} &
  \multicolumn{1}{c|}{} &
  \multicolumn{1}{c|}{} &
  \multicolumn{1}{c|}{20$\%$} &
  \multicolumn{1}{c|}{\textbf{1.01}} &
  \multicolumn{1}{c|}{\textbf{0.22}} &
  \multicolumn{1}{c|}{0.83} &
  \multicolumn{1}{c|}{0.23} &
  \multicolumn{1}{c|}{0.94} &
  0.39 \\ \cline{2-5} \cline{7-14} 
\multicolumn{1}{c|}{} &
  \multicolumn{1}{c|}{\multirow{4}{*}{0.1}} &
  \multicolumn{1}{c|}{5$\%$} &
  \multicolumn{1}{c|}{1.02} &
  \multicolumn{1}{c|}{0.24} &
  \multicolumn{1}{c|}{} &
  \multicolumn{1}{c|}{\multirow{4}{*}{0.3}} &
  \multicolumn{1}{c|}{5$\%$} &
  \multicolumn{1}{c|}{1.00} &
  \multicolumn{1}{c|}{0.38} &
  \multicolumn{1}{c|}{0.90} &
  \multicolumn{1}{c|}{0.44} &
  \multicolumn{1}{c|}{0.97} &
  0.59 \\ \cline{3-5} \cline{8-14} 
\multicolumn{1}{c|}{} &
  \multicolumn{1}{c|}{} &
  \multicolumn{1}{c|}{10$\%$} &
  \multicolumn{1}{c|}{1.02} &
  \multicolumn{1}{c|}{0.19} &
  \multicolumn{1}{c|}{} &
  \multicolumn{1}{c|}{} &
  \multicolumn{1}{c|}{10$\%$} &
  \multicolumn{1}{c|}{0.98} &
  \multicolumn{1}{c|}{0.36} &
  \multicolumn{1}{c|}{0.87} &
  \multicolumn{1}{c|}{0.33} &
  \multicolumn{1}{c|}{0.90} &
  0.48 \\ \cline{3-5} \cline{8-14} 
\multicolumn{1}{c|}{} &
  \multicolumn{1}{c|}{} &
  \multicolumn{1}{c|}{15$\%$} &
  \multicolumn{1}{c|}{\textbf{1.02}} &
  \multicolumn{1}{c|}{\textbf{0.18}} &
  \multicolumn{1}{c|}{} &
  \multicolumn{1}{c|}{} &
  \multicolumn{1}{c|}{15$\%$} &
  \multicolumn{1}{c|}{0.98} &
  \multicolumn{1}{c|}{0.28} &
  \multicolumn{1}{c|}{0.88} &
  \multicolumn{1}{c|}{0.31} &
  \multicolumn{1}{c|}{0.92} &
  0.42 \\ \cline{3-5} \cline{8-14} 
\multicolumn{1}{c|}{} &
  \multicolumn{1}{c|}{} &
  \multicolumn{1}{c|}{20$\%$} &
  \multicolumn{1}{c|}{1.02} &
  \multicolumn{1}{c|}{0.19} &
  \multicolumn{1}{c|}{} &
  \multicolumn{1}{c|}{} &
  \multicolumn{1}{c|}{20$\%$} &
  \multicolumn{1}{c|}{\textbf{1.02}} &
  \multicolumn{1}{c|}{\textbf{0.30}} &
  \multicolumn{1}{c|}{\textbf{0.87}} &
  \multicolumn{1}{c|}{\textbf{0.30}} &
  \multicolumn{1}{c|}{\textbf{0.94}} &
  \textbf{0.36} \\ \cline{2-5} \cline{7-14} 
\multicolumn{1}{c|}{} &
  \multicolumn{1}{c|}{\multirow{4}{*}{0.15}} &
  \multicolumn{1}{c|}{5$\%$} &
  \multicolumn{1}{c|}{1.02} &
  \multicolumn{1}{c|}{0.34} &
  \multicolumn{1}{c|}{} &
  \multicolumn{1}{c|}{\multirow{4}{*}{0.4}} &
  \multicolumn{1}{c|}{5$\%$} &
  \multicolumn{1}{c|}{1.02} &
  \multicolumn{1}{c|}{0.46} &
  \multicolumn{1}{c|}{0.88} &
  \multicolumn{1}{c|}{0.43} &
  \multicolumn{1}{c|}{0.94} &
  0.65 \\ \cline{3-5} \cline{8-14} 
\multicolumn{1}{c|}{} &
  \multicolumn{1}{c|}{} &
  \multicolumn{1}{c|}{10$\%$} &
  \multicolumn{1}{c|}{1.00} &
  \multicolumn{1}{c|}{0.29} &
  \multicolumn{1}{c|}{} &
  \multicolumn{1}{c|}{} &
  \multicolumn{1}{c|}{10$\%$} &
  \multicolumn{1}{c|}{\textbf{1.02}} &
  \multicolumn{1}{c|}{\textbf{0.39}} &
  \multicolumn{1}{c|}{0.84} &
  \multicolumn{1}{c|}{0.37} &
  \multicolumn{1}{c|}{0.94} &
  0.53 \\ \cline{3-5} \cline{8-14} 
\multicolumn{1}{c|}{} &
  \multicolumn{1}{c|}{} &
  \multicolumn{1}{c|}{15$\%$} &
  \multicolumn{1}{c|}{\textbf{1.02}} &
  \multicolumn{1}{c|}{\textbf{0.24}} &
  \multicolumn{1}{c|}{} &
  \multicolumn{1}{c|}{} &
  \multicolumn{1}{c|}{15$\%$} &
  \multicolumn{1}{c|}{1.06} &
  \multicolumn{1}{c|}{0.46} &
  \multicolumn{1}{c|}{\textbf{0.84}} &
  \multicolumn{1}{c|}{\textbf{0.39}} &
  \multicolumn{1}{c|}{0.93} &
  0.49 \\ \cline{3-5} \cline{8-14} 
\multicolumn{1}{c|}{} &
  \multicolumn{1}{c|}{} &
  \multicolumn{1}{c|}{20$\%$} &
  \multicolumn{1}{c|}{\textbf{1.01}} &
  \multicolumn{1}{c|}{\textbf{0.24}} &
  \multicolumn{1}{c|}{} &
  \multicolumn{1}{c|}{} &
  \multicolumn{1}{c|}{20$\%$} &
  \multicolumn{1}{c|}{\textbf{1.00}} &
  \multicolumn{1}{c|}{\textbf{0.39}} &
  \multicolumn{1}{c|}{0.85} &
  \multicolumn{1}{c|}{0.37} &
  \multicolumn{1}{c|}{\textbf{0.92}} &
  \textbf{0.48} \\ \cline{2-5} \cline{7-14} 
\multicolumn{1}{c|}{} &
  \multicolumn{1}{c|}{\multirow{4}{*}{0.2}} &
  \multicolumn{1}{c|}{5$\%$} &
  \multicolumn{1}{c|}{1.02} &
  \multicolumn{1}{c|}{0.33} &
  \multicolumn{1}{c|}{} &
  \multicolumn{1}{c|}{\multirow{4}{*}{0.5}} &
  \multicolumn{1}{c|}{5$\%$} &
  \multicolumn{1}{c|}{\textbf{0.98}} &
  \multicolumn{1}{c|}{\textbf{0.48}} &
  \multicolumn{1}{c|}{\textbf{0.90}} &
  \multicolumn{1}{c|}{\textbf{0.53}} &
  \multicolumn{1}{c|}{0.95} &
  0.74 \\ \cline{3-5} \cline{8-14} 
\multicolumn{1}{c|}{} &
  \multicolumn{1}{c|}{} &
  \multicolumn{1}{c|}{10$\%$} &
  \multicolumn{1}{c|}{1.03} &
  \multicolumn{1}{c|}{0.29} &
  \multicolumn{1}{c|}{} &
  \multicolumn{1}{c|}{} &
  \multicolumn{1}{c|}{10$\%$} &
  \multicolumn{1}{c|}{1.01} &
  \multicolumn{1}{c|}{0.47} &
  \multicolumn{1}{c|}{0.85} &
  \multicolumn{1}{c|}{0.44} &
  \multicolumn{1}{c|}{0.94} &
  0.63 \\ \cline{3-5} \cline{8-14} 
\multicolumn{1}{c|}{} &
  \multicolumn{1}{c|}{} &
  \multicolumn{1}{c|}{15$\%$} &
  \multicolumn{1}{c|}{1.01} &
  \multicolumn{1}{c|}{0.26} &
  \multicolumn{1}{c|}{} &
  \multicolumn{1}{c|}{} &
  \multicolumn{1}{c|}{15$\%$} &
  \multicolumn{1}{c|}{\textbf{0.99}} &
  \multicolumn{1}{c|}{\textbf{0.48}} &
  \multicolumn{1}{c|}{0.88} &
  \multicolumn{1}{c|}{0.44} &
  \multicolumn{1}{c|}{0.93} &
  0.62 \\ \cline{3-5} \cline{8-14} 
\multicolumn{1}{c|}{} &
  \multicolumn{1}{c|}{} &
  \multicolumn{1}{c|}{20$\%$} &
  \multicolumn{1}{c|}{\textbf{1.01}} &
  \multicolumn{1}{c|}{\textbf{0.25}} &
  \multicolumn{1}{c|}{} &
  \multicolumn{1}{c|}{} &
  \multicolumn{1}{c|}{20$\%$} &
  \multicolumn{1}{c|}{1.00} &
  \multicolumn{1}{c|}{0.46} &
  \multicolumn{1}{c|}{0.87} &
  \multicolumn{1}{c|}{0.46} &
  \multicolumn{1}{c|}{\textbf{1.02}} &
  \textbf{0.56} \\ \hline
\multicolumn{1}{c|}{\multirow{2}{*}{\textbf{DMBA$^{+}$}}} &
  \multicolumn{1}{c|}{1.2} &
  \multicolumn{1}{c|}{10$\%$} &
  \multicolumn{1}{c|}{\textbf{1.04}} &
  \multicolumn{1}{c|}{\textbf{1.12}} &
  \multicolumn{1}{c|}{\multirow{2}{*}{\textbf{DMBA$^{+}$}}} &
  \multicolumn{1}{c|}{1.2} &
  \multicolumn{1}{c|}{10$\%$} &
  \multicolumn{1}{c|}{\textbf{1.02}} &
  \multicolumn{1}{c|}{\textbf{1.13}} &
  \multicolumn{1}{c|}{0.98} &
  \multicolumn{1}{c|}{1.11} &
  \multicolumn{1}{c|}{\textbf{1.00}} &
   \textbf{1.15}\\ \cline{2-5} \cline{7-14} 
\multicolumn{1}{c|}{} &
  \multicolumn{1}{c|}{1.5} &
  \multicolumn{1}{c|}{10$\%$} &
  \multicolumn{1}{c|}{1.07} &
  \multicolumn{1}{c|}{1.38} &
  \multicolumn{1}{c|}{} &
  \multicolumn{1}{c|}{1.5} &
  \multicolumn{1}{c|}{10$\%$} &
  \multicolumn{1}{c|}{\textbf{1.03}} &
  \multicolumn{1}{c|}{\textbf{1.38}} &
  \multicolumn{1}{c|}{\textbf{1.02}} &
  \multicolumn{1}{c|}{\textbf{1.48}} &
  \multicolumn{1}{c|}{0.96} &
   1.35 \\ \hline
\multicolumn{1}{c|}{\multirow{2}{*}{\textbf{DMBA$^{++}$}}} &
  \multicolumn{1}{c|}{2} &
  \multicolumn{1}{c|}{10$\%$} &
  \multicolumn{1}{c|}{\textbf{1.08}} &
  \multicolumn{1}{c|}{\textbf{1.84}} &
  \multicolumn{1}{c|}{\multirow{2}{*}{\textbf{DMBA$^{++}$}}} &
  \multicolumn{1}{c|}{2} &
  \multicolumn{1}{c|}{10$\%$} &
  \multicolumn{1}{c|}{\textbf{1.00}} &
  \multicolumn{1}{c|}{\textbf{1.77}} &
  \multicolumn{1}{c|}{\textbf{1.01}} &
  \multicolumn{1}{c|}{\textbf{1.72}} &
  \multicolumn{1}{c|}{\textbf{0.97}} &
  \textbf{2.01} \\ \cline{2-5} \cline{7-14} 
\multicolumn{1}{c|}{} &
  \multicolumn{1}{c|}{3} &
  \multicolumn{1}{c|}{10$\%$} &
  \multicolumn{1}{c|}{1.10} &
  \multicolumn{1}{c|}{2.61} &
  \multicolumn{1}{c|}{} &
  \multicolumn{1}{c|}{3} &
  \multicolumn{1}{c|}{10$\%$} &
  \multicolumn{1}{c|}{1.00} &
  \multicolumn{1}{c|}{2.54} &
  \multicolumn{1}{c|}{1.05} &
  \multicolumn{1}{c|}{2.35} &
  \multicolumn{1}{c|}{0.98} &
  2.72 \\ \hline
\end{tabular}}

\caption{
\textbf{Attack performance of DMBAs against CSRnet on different datasets.} $\rho$ is the targeted manipulation ratio. Here, the poisoning rate $\gamma$ is fixed to 10\% while the trigger pattern is ``Rain". The most close-to-target results are \textbf{boldfaced}.
} 
\label{tab:across_dataset}
\vspace{-0.2cm}
\end{table*}
We first demonstrate the effectiveness of our attacks with the ``Rain" trigger pattern (see Figure \ref{fig:triggers}) from two perspectives: 1) the attack performance against the 5 counting models on the same dataset (i.e. SHB \cite{OoroRubio2016TowardsPO}); and 2) the attack performance against one most representative model (i.e. CSRNet) across different datasets. The $\hat{\rho}_{clean}$ and $\hat{\rho}_{dirty}$ results are reported in Table \ref{tab:across_model} and Table \ref{tab:across_dataset}, respectively, while the CMAE/AMAE and CRMSE/ARMSE results are in Appendix A.

\noindent\textbf{Effectiveness on Different Counting Models.} 
One key observation from Table \ref{tab:across_model} is that, against all 5 counting models, the backdoored models by our DMBAs have achieved a $\hat{\rho}_{dirty}$ that is very close to the backdoor target (the $\rho$ column) while maintaining a $\hat{\rho}_{clean}$ that is very close to 1. This verifies the effectiveness and stealthiness of our attacks on crowd counting models, i.e., the backdoored models have been successfully controlled to output the targeted counts (whether reduced or boosted) by the trigger pattern while behaving normally on clean test data.  
We can also observe that stronger backdoor targets (e.g., $\rho=0.2$ and $\rho=2$) are more difficult to achieve than weaker targets (e.g., $\rho=0.5$ and $\rho=1.2$), exhibiting larger gap from the targeted $\rho$. This is because, in crowd counting, learning exceptionally sparse or dense density maps can be overrode by the learning of regular density maps, making it hard to establish the backdoor correlation with extremely small or large $\rho$.
This phenomenon is different from classification backdoor attacks where the target is a hard label and the attack performance stays almost the same when choosing a different target \cite{nguyen2020input,li2021invisible}. This reveals one unique challenge of attacking regression models with respect to different targeted manipulation ratios.
In Table \ref{tab:across_model}, it also shows the failure ($\hat{\rho}_{dirty}$ remains close to 1) of TriOly (Trigger Only) attack which only injects the trigger pattern but does not alter the ground truth labels. This confirms the importance of density map altering for attacking crowd counting models.
By comparing the poisoning rates, we find that 15\%  poisoning is sufficient for effective attacks against the 5 models, and increasing the poisoning rate can further improve the attack but only slightly. 

\noindent\textbf{Effectiveness Across Different Datasets.} Here, we apply the ``Rain" trigger pattern with our DMBAs (poisoning rate $\gamma=10\%$) to attack the CSRnet \cite{Li2018CSRNetDC} on four different datasets. The results are reported in Table \ref{tab:across_dataset}.
At a high level, our attacks can consistently achieve the backdoor target (i.e., $\hat{\rho}_{dirty}$ is close to $\rho$) across the 4 datasets, with slight variations on different datasets.
This is because each dataset has its own learning difficulty (indicated by the closeness of $\hat{\rho}_{clean}$ to 1) and so is the attack. 
The SHA \cite{OoroRubio2016TowardsPO} dataset is notably harder to attack than the other 3 datasets, showing moderately larger gap between $\hat{\rho}_{dirty}$ and $\rho$, especially when $\rho$ is extremely small ($\rho=0.2$) or large ($\rho=2$). It also shows a larger $\hat{\rho}_{clean}$-to-1 and $\hat{\rho}_{dirty}$-to-$\rho$ gaps on SHA than on other datasets.
This indicates that datasets that consist of more diverse scenes and higher densities have certain natural robustness to backdoor attacks, a phenomenon that is also different to classification backdoor attacks where high attack success rates ($> 98\%$) can be easily achieved across different datasets \cite{li2021anti}.

By comparing the attack performance between DMBA$^{-}$ and DMBA$^{+}$/DMBA$^{++}$, we find that DMBA$^{-}$ is generally easier to achieve than DMBA$^{+}$/DMBA$^{++}$, a similar observation as in Table \ref{tab:across_model}. We conjecture this is because the task of learning low-density maps is easier than learning the high-density maps where the objects are highly overlapped.

\subsection{What Makes an Effective Trigger?}
\label{sec:understanding}
We conduct a set of ablation studies to help understand the key elements of effective crowd counting backdoor attacks. Here, we aim to gain understandings from two perspectives: 1) trigger type, and 2) trigger size. The experiments are run with CSRnet \cite{Li2018CSRNetDC} on SHA dataset \cite{OoroRubio2016TowardsPO}, the more challenging dataset with higher-density scenes. The backdoor target is set to $\rho=0.2$ and the poisoning rate is fixed to $\gamma$ = 20$\%$.
\begin{figure}
\vspace{-0.2cm}
	\centering
	\begin{minipage}{0.49\linewidth}
		\centering
		\includegraphics[width=0.9\linewidth]{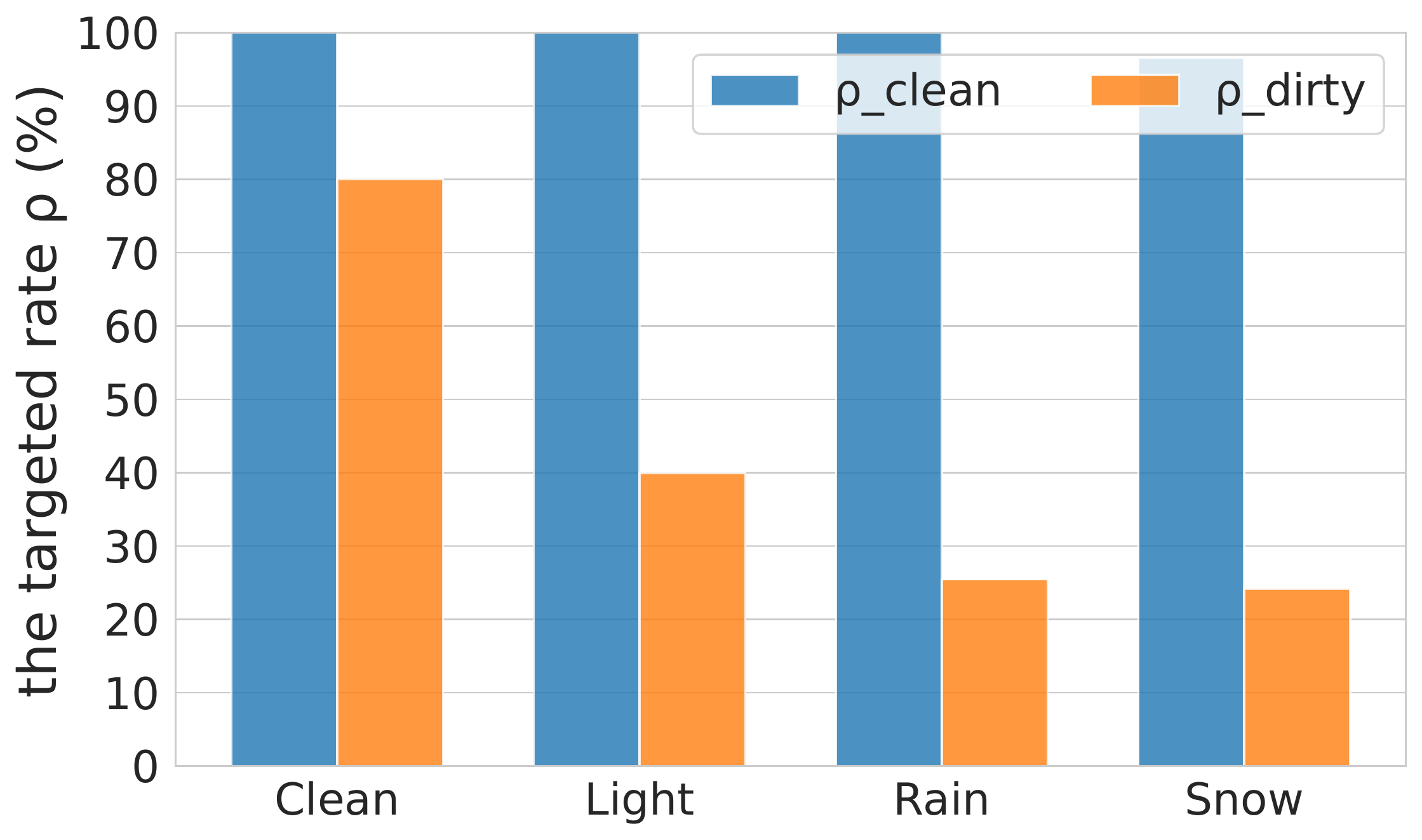}
		\label{CMSE/AMSE}
	\end{minipage}
	\begin{minipage}{0.49\linewidth}
		\centering
		\includegraphics[width=0.9\linewidth]{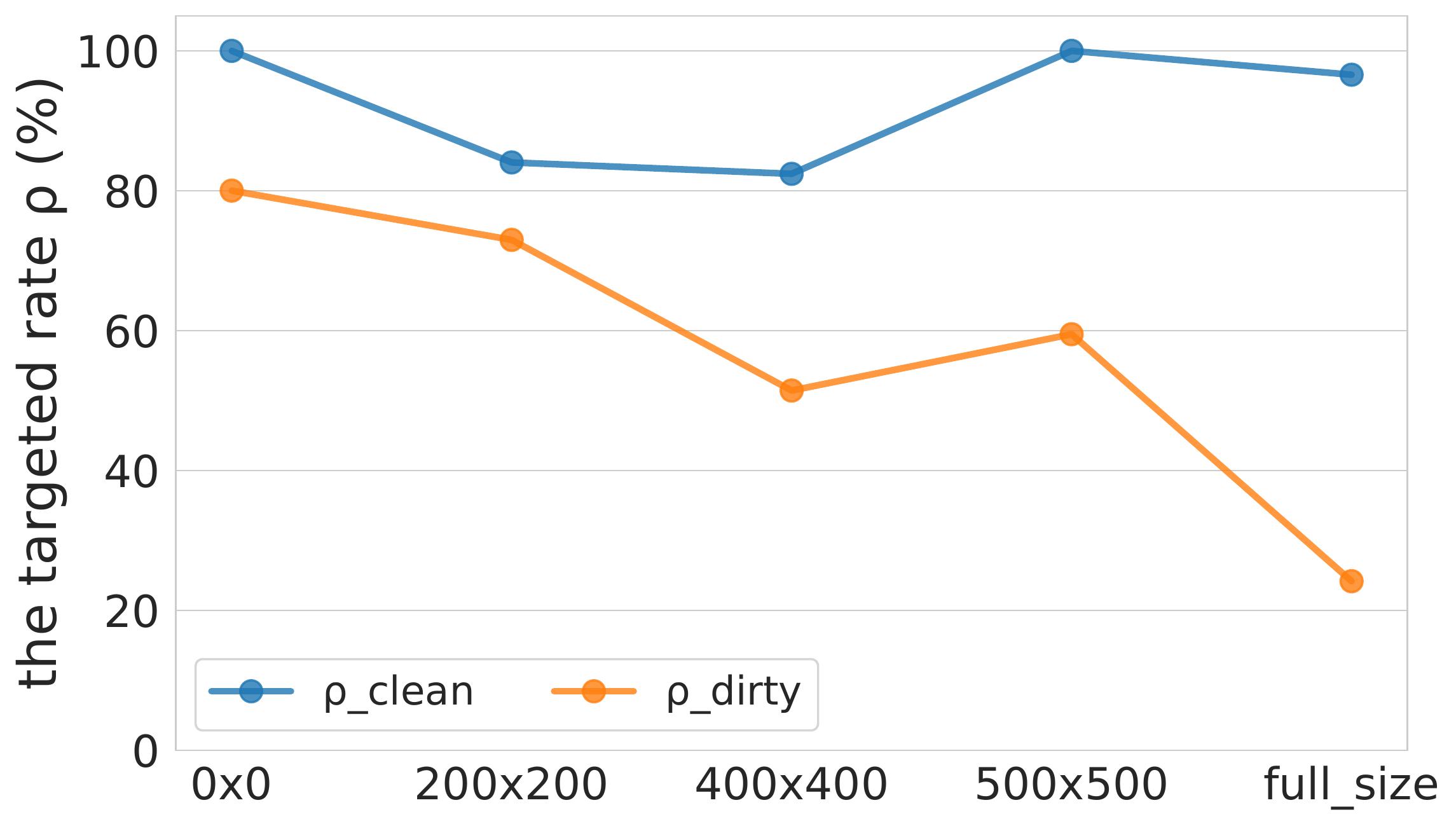}
		\label{CRMAE/ARMAE}
	\end{minipage}
\caption{Impact of Triggers. The experiments of triggers are conducted on CSRnet with SHA dataset, $\rho$ = 0.2 and $\gamma$ = 20$\%$.}
\label{fig:Impact of Triggers}
\vspace{-0.6cm}
\end{figure}

\noindent\textbf{Trigger type.}
We test the three types of trigger patterns visualized in Figure \ref{fig:triggers}: ``Rain", ``Snow" and ``Light". As shown in the left subfigure of Figure \ref{fig:Impact of Triggers}, ``Rain" and ``Snow" triggers are more effective than the ``Light" trigger, demonstrating closer $\hat{\rho}_{dirty}$ to 0.2 which is the attack target. This is because , compared with ``Light", ``Rain" and ``Snow" have more scattered points that greatly interfere with the heads/objects in the original image. This experiment reveals the importance of using densely scattered trigger patterns to attack crowd counting models. Note that the trigger patterns are not restricted to natural effects, although we believe such triggers can make the attack more stealthy and are easy to simulate in real-world attacks.

\noindent\textbf{Trigger size.} 
Here, we take the ``Snow" trigger as an example and test 5 different sizes including 0x0, 200x200, 400x400, 500x500, and the same size of the input image (see Figure \ref{fig:Trigger working}). The results are presented in the right subfigure of Figure \ref{fig:Impact of Triggers}. It is evident that larger trigger patterns have a clear advantage in achieving the backdoor target $\rho=0.2$ than small trigger patterns. This verifies the importance of using large trigger patterns to force the model to learn the backdoor correlation. Again, this is because learning the backdoor in crowd counting models is not a (relatively) independent task but rather a highly-interfered process with the original density regression task. Combining our finding in the previous experiment, we conclude that large and densely scattered background trigger patterns are the key to effective crowd counting backdoor attacks.

\subsection{Resistance to Advanced Backdoor Defenses}\label{sec:resistance}
Here, we test the effectiveness of three (two classic and one advanced) backdoor defense methods developed for classification models against our DMBA$^{-}$ attack: 1) Pruning \cite{Dhillon2018StochasticAP}, 2) Fine-Pruning \cite{liu2018fine-pruning}, and 3) Adversarial Neural Pruning (ANP) \cite{wu2021adversarial}. We run these experiments with CSRnet model on SHA dataset with trigger pattern ``Rain", backdoor target $\rho=0.2$ and poisoning rate $\gamma=10\%$.

\noindent\textbf{Pruning and Fine-Pruning.} 
Both methods prune the neurons that stay dormant on clean inputs but can potentially be triggered by backdoored inputs to mitigate backdoor attacks. 
Here, we apply pruning and fine-pruning on the DMBA$^{-}$-backdoored CSRnet to prune neurons from the last two layers of both the back-end and front-end modules of the network. We follow the \textit{prune-then-test} pipeline for pruning and the \textit{prune-finetune-then-test} pipeline for Fine-pruning. We prune the the selected layers until 90\% of the neurons are removed. We find that both defenses can mitigate our DMBA$^{-}$ attack to certain extent, yet they both significantly degrade the model's performance on the clean inputs. Particularly, when $\hat{\rho}_{dirty}$ was recovered to 0.95 (originally close to the backdoor target 0.2), the $\hat{\rho}_{dirty}$ of the model increases to 1.83 (originally close to 1) while the MAE increases from 92.41 to 415.55.
More detailed results can be found in Appendix B.

\noindent\textbf{Adversarial Neuron Pruning (ANP).}
ANP is one of the state-of-the-art defense methods that locates and prunes backdoor neurons based on the neurons' sensitivity to adversarial perturbations \cite{wu2021adversarial}. 
ANP was originally proposed for classification models. Here we adapt ANP to defend crowd counting models. We first optimized the neural perturbation as follows:
\begin{equation}\label{eq4}
    \max _{\boldsymbol{\delta}, \boldsymbol{\xi} \in [-\epsilon, \epsilon]^n} \mathcal{L}_{\mathcal{D}_v}=\frac{1}{2 N_{v}}\left\|\hat{z}_{i}-z_{i}^{g t}\right\|_{2}^{2},
\end{equation}
where, $\boldsymbol{\delta}$, $\boldsymbol{\xi}$ and $\epsilon$ are the 3 hyper-parameters of ANP \cite{wu2021adversarial}, $n$ is the number of neurons, and
$N_{v}$ is the number of clean samples from a small clean validation set $\mathcal{D}_v$. 
ANP then locates backdoor neurons by learning a mask via the following:
\begin{equation}
    \min _{\bold{m} \in [0,1]^{n}}\left[\alpha \mathcal{L}_{\mathcal{D}_v}(\bold{m} \odot \theta)+(1-\alpha) \cdot \max _{\boldsymbol{\delta}, \boldsymbol{\xi}} \mathcal{L}_{\mathcal{D}_v}([\bold{\delta}, \bold{\xi}] \odot \theta)\right],
\end{equation}
where, $\theta$ are the model parameters, $\bold{m}$ is the neuron mask and $\alpha$ is a trade-off coefficient \cite{wu2021adversarial}.
After 2000 iterations of pruning following the setting in \cite{wu2021adversarial}, $\hat{\rho}_{dirty}$ of the backdoored CSRnet model is still 0.25, which is close to no ANP defense ($\hat{\rho}_{dirty}=0.21$). 
 Moreover, the results on the clean test data show that ANP greatly degrade the model's clean performance (see Figure \ref{fig:prediction after ANP defense} in Appendix B). 

The above results indicate that backdoored neurons are mixed with clean neurons in the backdoored models by our DMBA$^{-}$ attack, making it hard to segregate and prune backdoor neurons without hurting the clean performance. This means that our DMBA$^{-}$ attack is fairly resistant to these defense methods.

\section{Conclusion}
In this paper, we studied the problem of backdoor attack on crowd counting models. We first verified the ineffectiveness of classification backdoor attacks on crowd counting models, then proposed two novel Density Manipulation Backdoor Attacks (DMBA$^{-}$ and DMBA$^{+}$) to attack crowd counting models to produce targeted count estimations. We demonstrated the effectiveness of our DMBA attacks on 5 crowd counting models and 4 datasets with low poisoning rate. We also provide an analysis of the key elements of effective attacks: 1) large and dense trigger patterns, and 2) the alteration of the ground truth density maps. We hope our attacks can serve as strong baselines for the robustness evaluation of crowd counting models and development of effective defenses \cite{Zhu2020FreeLB,chen2020adversarial}.

\section{Acknowledgments}

This work is supported by National Natural Science Foundation of China (NSFC) under grant no. 61972448. (Corresponding author: Pan Zhou). We would like to thank all anonymous reviewers for their constructive feedback.

\clearpage
\bibliographystyle{ACM-Reference-Format}
\bibliography{sample-authordraft}

\clearpage
\appendix

\end{document}